\documentclass[twocolumn]{IEEEtran}

\usepackage[pdftex]{graphicx} % 插图命令 \includegraphics
\usepackage{amsmath} % Math Packages
\usepackage{algorithmic}
\usepackage{array} % Alignment Packages
\usepackage[caption=false,font=footnotesize]{subfig} % Subfigure Packages
\usepackage{dblfloatfix}
\usepackage{url} % simple URL typesetting

\usepackage{microtype}
\usepackage[utf8]{inputenc} % allow utf-8 input
\usepackage[T1]{fontenc}    % use 8-bit T1 fonts
\usepackage{paralist} % Compacted versions of enumerate and itemize.
\usepackage{marvosym} % 常用符号, 特别是打勾、打叉
\usepackage{bm} % 实现字体的斜体加粗
\usepackage{bbm} % fix \mathbb doesn't support digits
\usepackage{amssymb}   % 黑板粗体 \mathbb, 德文尖角体 \mathfrak, \backsim
\usepackage{mathrsfs} % 花体 \mathscr
\usepackage{tabularray} % tblr or talltblr or longtblr environment
\UseTblrLibrary{booktabs}       % professional-quality tables，也就是三线表
\usepackage{tabularx} % 自动计算列宽的表格包
\usepackage[flushleft]{threeparttable} % 给表格项添加注释
\usepackage{makecell} % 表格内容换行
\usepackage{multirow} % 多行表格
\usepackage{nicefrac} % 斜着的分号
\usepackage[dvipsnames]{xcolor} % 字体颜色
\usepackage{soul} % highlight, strikeout, underline
\setul{0.5ex}{0.3ex}

\definecolor{citecolor}{RGB}{34,139,34}
\usepackage[pagebackref=true,breaklinks=true,letterpaper=true,colorlinks,
    citecolor=citecolor,bookmarks=false]{hyperref}
\usepackage{cleveref}  % for \cref
\crefformat{figure}{Fig.~#2#1#3}
\crefformat{equation}{Eq.~(#2#1#3)}
\crefformat{table}{Table~#2#1#3}
\crefformat{section}{Section~#2#1#3}
\crefformat{algorithm}{Algorithm~#2#1#3}
\crefformat{appendix}{Appendix #2#1#3}
\crefformat{subsection}{subsection~#2#1#3}
\usepackage{cellspace}
\usepackage[numbers,sort&compress]{natbib} % 代替 cite.sty

\usepackage{xspace} % 用于在 \newcommand 中添加空格

\usepackage{pifont} % http://ctan.org/pkg/pifont
\newcommand{\xmark}{\ding{53}}%

\begin{document}

\setlength{\abovedisplayskip}{.5\baselineskip} % 调整公式与正文间的段前距离
\setlength{\belowdisplayskip}{.5\baselineskip} % 调整公式与正文间的段后距离

% OSCAR
\title{One-Stage Cascade Refinement Networks for Infrared Small Target Detection}

% !TEX root = ../main.tex

\author{
  Yimian~Dai,
  Xiang~Li,
  Fei~Zhou,
  Yulei~Qian,
  Yaohong~Chen,
  Jian~Yang
  \thanks{
    % Manuscript received XXXX YY, 2022; revised XXXX YY, 2022; accepted XXXX YY, 2022.
    This work was supported by
    % 博后面上
    the fellowship of China Postdoctoral Science Foundation (No. 2021M701727), the National Key R\&D Plan of the Ministry of Science and Technology (Project No.2020AAA0104400), the Young Scientists Fund of the National Natural Science Foundation of China (Grant No.62206134), and the National Science Fund of China (Grant No.U1713208).
    \emph{(Corresponding author: Xiang Li and Jian Yang.)}
    }

  % 南理工
  \thanks{
    Yimian Dai and Jian Yang are with PCA Lab, Key Lab of Intelligent Perception and Systems for High-Dimensional Information of Ministry of Education, School of Computer Science and Engineering, Nanjing University of Science and Technology, Nanjing, China.
    (e-mail:
    \href{mailto:yimian.dai@gmail.com}{yimian.dai@gmail.com};
    \href{mailto:csjyang@mail.njust.edu.cn}{csjyang@mail.njust.edu.cn}).
  }
  
  \thanks{Xiang Li is with IMPlus@PCA Lab, College of Computer Science, Nankai University. (e-mail:\href{mailto:xiang.li.implus@nankai.edu.cn}{xiang.li.implus@nankai.edu.cn}).
  }

  % 河工大
  \thanks{Fei Zhou is with College of Information Science and Engineering, Henan University of Technology, Zhengzhou, China.
  (e-mail: \href{mailto:hellozf1990@163.com}{hellozf1990@163.com}).
  }

  % 724
  \thanks{Yulei Qian is with Nanjing Marine Radar Institute, Nanjing, China
  (e-mail: \href{mailto:yuleifly@126.com}{yuleifly@126.com}).
  }

  % 光机所
  \thanks{
    Yaohong Chen is with Xi'an Institute of Optics and Precision Mechanics, Chinese Academy of Sciences, Xi'an, China.
    (e-mail: \href{mailto:chenyaohong@opt.ac.cn}{chenyaohong@opt.ac.cn}).
  }
}

\maketitle

% !TEX root = ../main.tex

\begin{abstract}
Single-frame InfraRed Small Target (SIRST) detection has been a challenging task due to a lack of inherent characteristics, imprecise bounding box regression, a scarcity of real-world datasets, and sensitive localization evaluation. In this paper, we propose a comprehensive solution to these challenges. First, we find that the existing anchor-free label assignment method is prone to mislabeling small targets as background, leading to their omission by detectors. To overcome this issue, we propose an all-scale pseudo-box-based label assignment scheme that relaxes the constraints on scale and decouples the spatial assignment from the size of the ground-truth target. Second, motivated by the structured prior of feature pyramids, we introduce the one-stage cascade refinement network (OSCAR), which uses the high-level head as soft proposals for the low-level refinement head. This allows OSCAR to process the same target in a cascade coarse-to-fine manner. Finally, we present a new research benchmark for infrared small target detection, consisting of the SIRST-V2 dataset of real-world, high-resolution single-frame targets, the normalized contrast evaluation metric, and the DeepInfrared toolkit for detection. We conduct extensive ablation studies to evaluate the components of OSCAR and compare its performance to state-of-the-art model-driven and data-driven methods on the SIRST-V2 benchmark. Our results demonstrate that a top-down cascade refinement framework can improve the accuracy of infrared small target detection without sacrificing efficiency. The DeepInfrared toolkit, dataset, and trained models are available at \url{https://github.com/YimianDai/open-deepinfrared}.
%  to advance further research in this field.
\end{abstract}

\vspace{-1\baselineskip}
% !TEX root = ../main.tex
% \bibliography{../../../ref-bib/all-refs.bib}

\section{Introduction}  \label{sec:introduction}
% 1. 研究背景的意义
% 2. 国内外发展动态分析
% 3. 关键技术瓶颈/关键科学问题
% 4. 研究目标/文章动机
% 5. 技术路线
% 6. 本研究的意义
% 7. 章节安排

% 1. 研究背景的意义
% 红外成像系统是重要的载荷
Infrared imaging systems have become critical optoelectronic components in surveillance systems due to their ability to detect objects at any time of day, their long range, and their ability to capture heat sources \cite{GRMS2022InfraredSurvey}.
% 红外小目标检测是关键任务
Long-distance targets of interest often appear as small targets in the field of vision because of the requirements of early warning detection systems and the resolution of infrared imaging systems.
As a result, infrared small target detection is essential for many applications and has been a focus of research in recent years \cite{PR21RingTopHat}.

% 主要都采用语义分割的形式
Since the seminal work of Miss Detection vs. False Alarm (MDvsFA) \cite{ICCV19Infrared} and Asymmetric Contextual Modulation Networks (ACMNet) \cite{WACV21ACM}, many Convolutional Neural Network (CNN) based methods have been proposed \cite{TGRS2021ALCNet} and model the infrared small target detection as a semantic segmentation problem \cite{CVPR15FCN}. Despite their success, it has been difficult to further improve the detection performance even with deeper backbones \cite{CVPR16ResNetV1} and more powerful attention modules \cite{CVPR19DualAttention}.

% 语义分割存在的问题
We argue that the performance bottleneck in infrared small target detection originates in the way the task is modeled as semantic segmentation\cite{TCSVT21TensorFieldGraphCut}, which is unreasonable for infrared small targets in three ways.
% 1) 只是中间表示
First, for real-world applications, semantic segmentation is only an intermediary representation; the ultimate objective is to locate the centroids of infrared small targets so they can be used as input for the tracking task. Therefore, segmentation integrity is only a rough approximation of detection accuracy.
% 2) 标记存在模糊, 且这种模糊会 dominate positive pixels
Second, due to the dispersion effect of long-distance imaging, it is impossible for humans to label infrared small targets with pixel-level accuracy, and ground-truth annotations include inherent ambiguities. The impact of these uncertain pixels is significant because they make up the majority of positive samples and their loss dominates the training process.
% 3) 计算量很大
Third, in order to segment infrared small targets accurately, inference must be performed on high-resolution feature maps, making it computationally intensive. Therefore, considering both detection accuracy and computational overhead, it is better to model infrared small target detection as an object detection problem rather than a semantic segmentation problem.
However, the performance of infrared small target detection still lags behind that of normal-scale object detection by a significant margin.

The performance gap in infrared small target detection is mainly caused by three factors.
% 标签分配
First, current label assignment paradigms, whether anchor-based or anchor-free, tend to mistakenly label tiny groundtruth targets as background, leading detectors to pay less attention to small targets.
% 定位不准
Second, for infrared small targets, a small perturbation in the bounding box can cause a significant disturbance in the Intersection over Union (IoU) metric, making both bounding box regression and localization evaluation more challenging than for large objects.
% 缺少数据
Third, collecting real-world infrared small target images is more challenging than collecting generic images, resulting in a lack of non-synthetic data to train machine learning models.

\begin{figure*}[htbp]
  \centering
  \subfloat[Faster RCNN]{
    \includegraphics[width=.46\textwidth]{
      "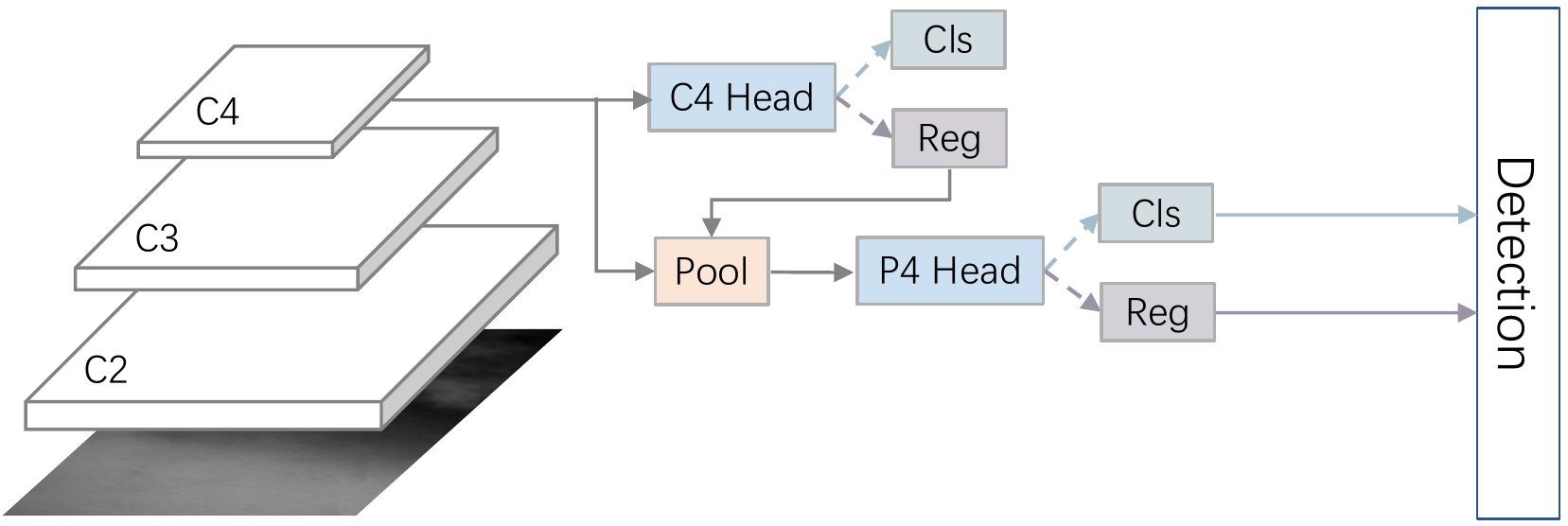"}
  }
  \subfloat[Cascade RCNN]{
    \includegraphics[width=.46\textwidth]{
      "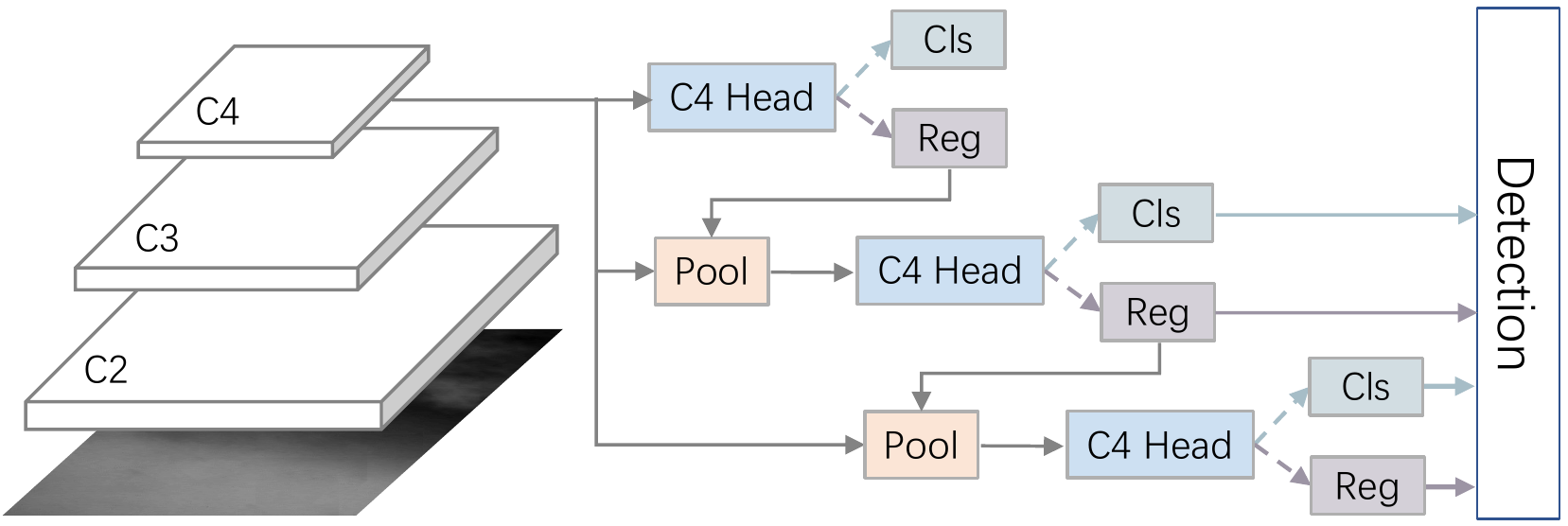"}
      % \label{subfig:plot-neg-pos-ratio}
  }

  \subfloat[RefineDet]{
    \includegraphics[width=.46\textwidth]{
      "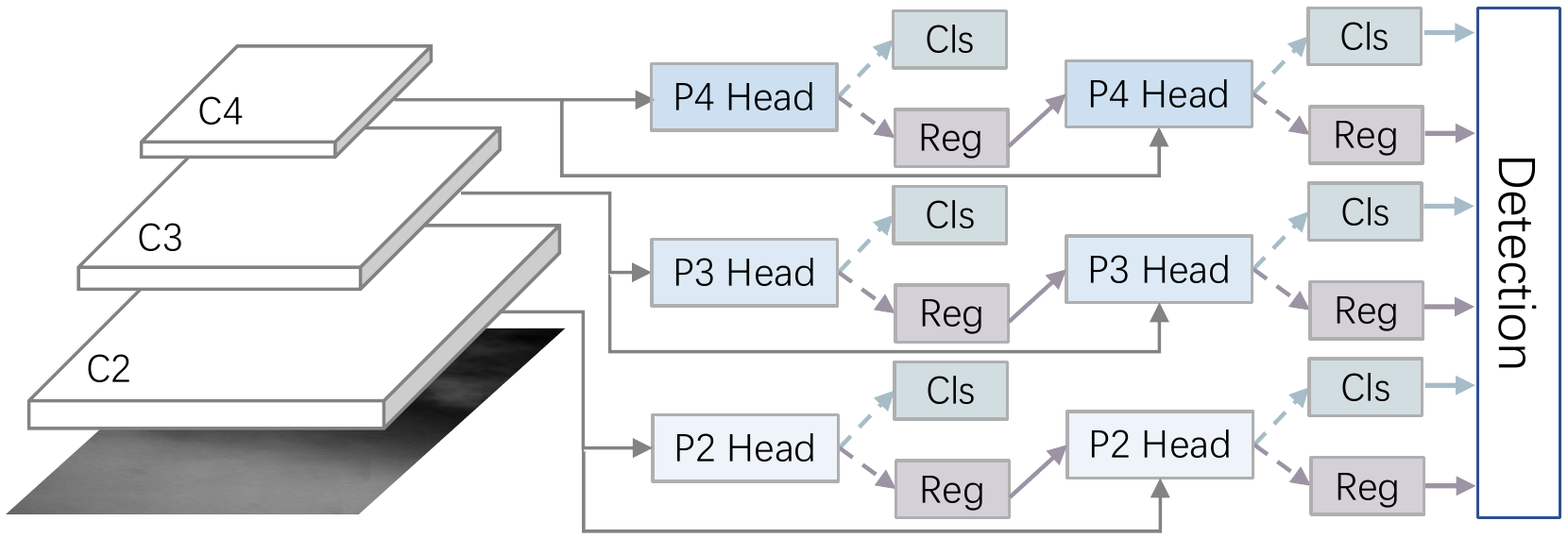"}
      % \label{subfig:plot-compute-load}
  }
  \subfloat[OSCAR]{
    \includegraphics[width=.46\textwidth]{
      "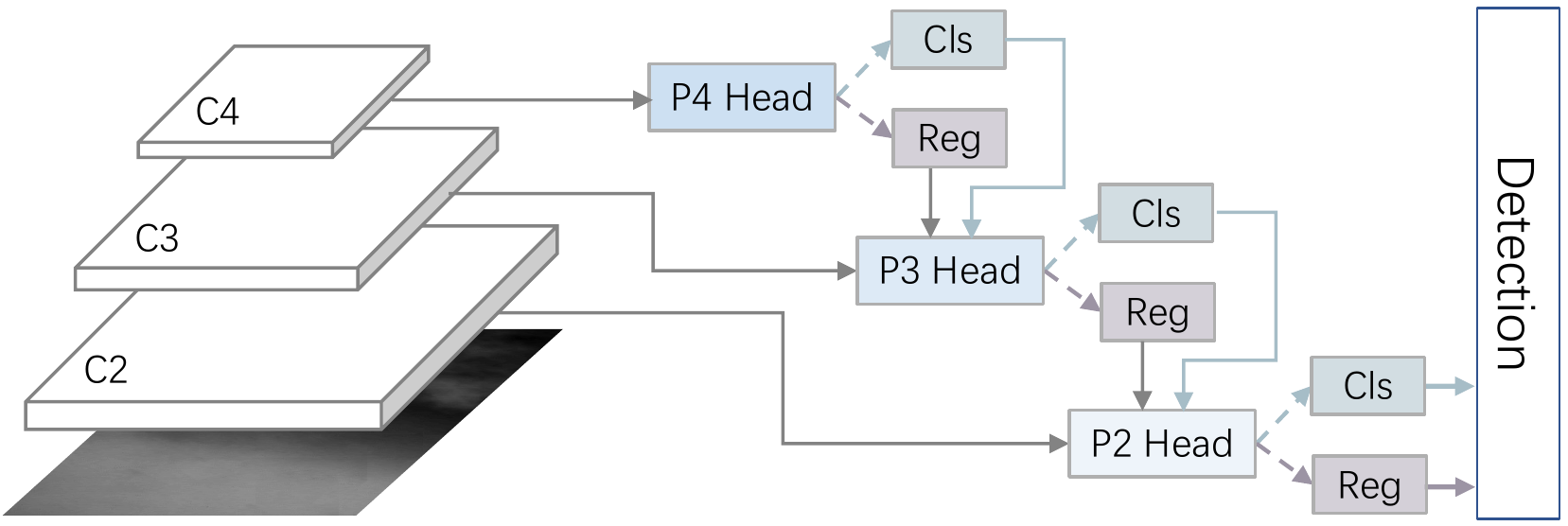"}
      % \label{subfig:plot-compute-load}
  }
  \caption{
    Comparisons of architecture frameworks for cascade detection between existing works and ours.
    Existing works focus on cascade detection the object on the same level of feature map, including (a): Faster RCNN \cite{NIPS15FasterRCNN}, which detects objects in two stages on C2 layer, (b) Cascade RCNN \cite{CVPR18CascadeRCNN}, which detects objects in multiple stages on C2 layer, (c): RefineDet \cite{CVPR18RefineDet}, which parallelly detects the object on each feature maps of different scales in a cascaded manner.
    In contrast, our OSCAR approach utilizes the highly structured feature pyramids to perform cascaded detection in a top-down, coarse-to-fine manner.
    Here, "Pool" denotes region-wise feature extraction, "Cls" indicates the classification, "Reg" stands for the bounding box regression.
    The backbone feature levels are denoted as "C2", "C3", and "C4", while the levels of the feature pyramids are denoted as "P2", "P3", and "P4".}
  \label{fig:abbr-arch}
  \vspace{-1\baselineskip}
\end{figure*}

In this paper, we propose a comprehensive solution to address the challenges faced in infrared small target detection. We first introduce a new research benchmark, consisting of (1) SIRST-V2, a dataset of real-world, high-resolution single-frame infrared small targets, (2) normalized contrast, a reliable measure of localization quality that accounts for the inherent label ambiguity in long-range imaging, and (3) DeepInfrared, an open source toolkit specifically designed for infrared small target detection. We hope that our benchmark and toolkit will facilitate the development of novel algorithms for this task.

To address the performance gap in infrared small target detection, we also propose a new label assignment scheme called All-Scale Pseudo-Box (ASPB). This scheme decouples the ground-truth target size from the spatial assignment by using scale-adaptive pseudo-boxes and also relaxes the scale constraints by treating all target boxes at all scales as positive samples.
This new scheme is motivated by the observation that infrared small targets, such as those with bounding boxes of $3\times3 \sim 5\times5$, may not contain any feature points when using a common feature map stride, such as $8$. This is a notable difference from generic object detection.
However, infrared small target detection is actually a sparse target detection task with an occlusion-free prior and a known scale. In the highly structured feature pyramids \cite{CVPR17FPN}, the multi-scale predictions can be viewed as multiple approximations of the target in a coarse-to-fine manner, which allows for more accurate localization.
% This allows for more accurate localization and better handling of the label ambiguities caused by long-range imaging.

Finally, motivated by this cross-scale spatial correspondence, we propose the \textbf{O}ne-\textbf{S}tage \textbf{Ca}scade \textbf{R}efinement (OSCAR) network, a new data-driven framework for single-frame infrared small target detection that aims to perform cascaded bounding box regression while still maintaining the efficiency of one-stage methods.
In contrast to generic object detection, which adopts a divide-and-conquer strategy to represent different scales of objectives with different layers, OSCAR reformulates a feature pyramid's multi-level predictions as approximations of cascaded soft region proposals in a coarse-to-fine order towards the same ground-truth bounding box.
To address the class imbalance issue and down-weigh well-classified negative samples, a top-down cross-head modulation mechanism is used as a soft region proposal module.
We also incorporate a normalized contrast branch in the refining head, which significantly improves performance by suppressing low-quality predicted bounding boxes caused by pseudo-boxes.

Our main contributions in this work are:
\begin{enumerate}
  \item We propose a real-world high-resolution dataset for infrared small target detection, called \emph{SIRST-V2}.
  % \item We propose a real-world high-resolution dataset for infrared small target detection, called the \emph{SIRST-V2} benchmark dataset.
  % To the best of our knowledge, this is the largest non-synthetic dataset in this field.
  \item We design \emph{normalized contrast}, a novel evaluation metric for infrared small targets, which takes into account centroid localization and robustness under the ambiguities of long-range imaging.
  \item We provide an open-source toolkit, called \emph{DeepInfrared}, to support researchers in developing algorithms for the SIRST-V2 benchmark and normalized contrast metric.
  % This toolkit also includes implementations and trained models of many well-known algorithms for infrared small target detection.
  \item We propose \emph{OSCAR}, a new framework for single-frame infrared small target detection that aims to perform cascaded bounding box regression in a top-down cascade manner. By utilizing the conclusion-free prior in highly structured feature pyramids, our method achieves state-of-the-art detection performance while maintaining the efficiency of one-stage methods.
\end{enumerate}
\vspace{-.5\baselineskip}
% To verify the effectiveness of the proposed OSCAR, we conduct extensive ablation studies to investigate the importance of pseudo-box-based label assignment as well as each component of the OSCAR head, including top-down score modulation module, cascade bounding box refinement, and the normalized contrast branch.
% We also compare it with other state-of-the-art model-driven methods and data-driven methods on the public SIRSTv2 benchmark. The experimental results indicate that the proposed OSCAR achieves the best performance.

% !TEX root = ../main.tex
% \bibliography{../../../ref-bib/all-refs.bib}

\section{Related Work} \label{sec:related-work}

\subsection{Infrared Small Target Detection}
\subsubsection{Traditional Methods}
% 传统方法
Infrared small target detection is typically formulated as either a target enhancement problem \cite{TGRS20TopHatReg,TAES20RingTopHat} or a background suppression problem \cite{GRSL2019LocalContrastTDLMS}. These methods can be further classified based on the modeling assumptions they use, such as local contrast approaches \cite{TGRS13LCM} and sparse plus low-rank decomposition methods \cite{TIP13IPI,TGRS22ThreeOrderTensor}.
% local contrast method 是什么
Local contrast methods model infrared small targets as outliers in a slowly-changing background of highly correlated pixels \cite{TGRS13LCM}. Different variants of local contrast measures have been proposed to enhance small targets and suppress pixel pulse noise and edges simultaneously \cite{GRSL22WeightedThreeLayerWindow}.
% low-rank 方法是什么
Sparse plus low-rank decomposition methods exploit the non-local self-correlation property of background patches by assuming that all background patches come from a single subspace or a mixture of low-rank subspace clusters \cite{TIP13IPI,JSTARS17RIPT}. This allows target-background separation to be achieved by decomposing the original data into a sparse target image and a low-rank background image \cite{TIP20TNLRS}.

% 缺点是什么
However, in real-world scenarios, there are many distractors that stand out as outliers in the background alongside true targets \cite{IPT16WIPI}. Traditional approaches lack the ability to differentiate between true targets and similar distractors at a high semantic level, because they only use grayscale values or low-level handcrafted features such as central differences and entropy \cite{TGRS2021ALCNet}. When given an infrared image without any targets, traditional approaches often yield false positive detections because the contrasts they predict are not semantically normalized.
Another problem is that the contrast values of different targets in multi-target settings cannot be normalized to the same scale, which makes it more likely to miss targets with low contrast values and results in a high number of missed detections.
These modeling issues can cause traditional methods to perform poorly when dealing with complicated scenarios.
% In practical applications, these modeling issues can cause traditional methods to perform poorly when dealing with complicated scenarios.

\subsubsection{Deep Learning Methods}
% 深度学习方法
Infrared small target detection can now be modeled as a supervised machine-learning problem due to the availability of publicly accessible datasets like SIRST-V1 \cite{WACV21ACM}.
Currently, most studies focus on designing multi-scale feature fusion models that can produce more refined target representations because of the problem of insufficient intrinsic features \cite{CVPR17FPN}. To exchange high-level semantics and low-level details, Dai \textit{et al.} proposed an asymmetric contextual modulation module that incorporates a bottom-up modulation pathway based on point-wise channel attention and a top-down global contextual feedback \cite{WACV21ACM}.
To enhance features that were matched with prior knowledge, Dai \textit{et al.} proposed a specialized nonlinear feature transformation layer, which they embedded into a deep network \cite{TGRS2021ALCNet}. Meanwhile, Li \textit{et al.} designed an interactive module that is densely nested, allowing for gradual interaction between high-level and low-level elements \cite{arXiv21DNANet}.
Another significant challenge in infrared small target detection is achieving a balance between miss detections and false alarms. To address this, Wang \textit{et al.} proposed splitting the task into two smaller tasks, each of which would be handled by an adversarially trained model that focuses on reducing either false alarms or miss detections \cite{ICCV19Infrared}.

% Zhao et al. modeled the detection problem as a translation problem between raw images to object-only detection images via a U-Net \cite{MICCAI15UNet} based generative adversarial network \cite{TGRS21InfraredGAN}.

Although these methods significantly outperform traditional methods, formulating the detection problem as a semantic segmentation or image translation \cite{TGRS21InfraredGAN} requires computationally expensive high-resolution feature maps to recover pixel-level details.
In contrast to the pixel-wise segmentation methods mentioned above, we formulate the infrared small target detection as a bounding box regression task, which is less affected by labeling ambiguity.
Furthermore, in the proposed OSCAR, the top-down modulation is applied to the detection head on the label space rather than the feature space in the neck, specifically the feature pyramid \cite{CVPR17FPN}.

\subsection{Generic Object Detection}
% Object detection is one of the fundamental tasks for computer vision applications. Recent years have witnessed a dramatic improvement in object detection.
% To some extent, infrared small target detection can be regarded as a special case of detecting extremely small-scale objects under the generic object detection framework.

\subsubsection{Two-Stage Methods}

The two-stage detector, such as Faster R-CNN \cite{NIPS15FasterRCNN}, is a paradigm of object detection algorithm that first generates a set of region proposals and then classifies the objects within those proposals (see \cref{fig:abbr-arch} (a)).
% To some extent, two-stage object detection can be thought of as a cascade, where the first detector removes a significant number of background samples and the second stage classifies the remaining regions.
% Furthermore, Cascade R-CNN improves two-stage methods by using cascade sub-networks to gradually improve the quality of region proposals \cite{CVPR18CascadeRCNN}.
These algorithms can be conceptualized as cascades, where the first stage removes a significant number of background samples and the second stage classifies the remaining regions. In addition, Cascade R-CNN \cite{CVPR18CascadeRCNN} extends the two-stage paradigm by using multiple sub-networks to progressively improve the quality of the region proposals (see \cref{fig:abbr-arch} (b)).
Our OSCAR follows the idea of a cascade refinement, but it differs in two important aspects:
(1) Instead of using a sparse set of candidate object boxes for the next stage, OSCAR performs soft region proposal on dense predictions by down-weighting easy background samples;
(2) Although it performs progressive refinement on the predicted results, OSCAR remains a one-stage detector, not a two-stage method, avoiding the additional computational overhead of general cascading pipelines.

% Faster R-CNN , a representative of two-stage techniques, has become the dominant paradigm in the modern era of object detection, outperforming other paradigms on numerous benchmarks such as IJCV 2010 VOC and ECCV 2014 COCO \cite{IJCV2010VOC,ECCV14COCO}. For instance, even though it is orders of magnitude faster than Fast RCNN \cite{ICCV15FastRCNN}, the region-wise sub-network of Faster RCNN still needs to be applied to every region of interest (RoI), and an image typically has several hundred RoIs.
% To address this issue, Dai \textit{et al.} introduced the Region-based Fully Convolutional Network (RFCN) \cite{NIPS16RFCN}, which builds a shared RoI sub-network using all convolution layers and extracts RoI crops from the final layer features before prediction."
% Additionally, two-stage object detection can be thought of as a cascade, where the first detector removes a significant number of background samples and the second stage classifies the remaining regions.
% Furthermore, Cascade R-CNN improves two-stage methods by using cascade sub-networks to gradually improve the quality of region proposals \cite{CVPR18CascadeRCNN}.

\subsubsection{Single-Stage Methods}
One-stage approaches, such as DenseBox \cite{arXiv15DenseBox} and CenterNet \cite{arXiv19CenterNetKeypoint}, bypass the region proposal step and directly predict classification scores and bounding box regression offsets, aiming for real-time performance while maintaining high accuracy.
% However, these methods still require densely distributed anchors, which are produced by sliding windows \cite{ECCV16SSD}. This can be problematic because the extreme foreground-background class imbalance can limit the performance of the method.
% To address this, some one-stage methods use techniques such as focal loss \cite{ICCV17FocalLoss} to rectify the Cross Entropy loss by down-weighting the loss assigned to correctly classified examples.
There are some works that try to transplant the merits of two-stage methods to one-stage approaches. For example, RefineDet \cite{CVPR18RefineDet} uses a refinement module to mimic the second regression of two-stage approaches (see \cref{fig:abbr-arch} (c)), and Consistent Optimization \cite{arXiv19ConsistentOptim} reduces the gap between the training and testing phases by tying consecutive classification targets to the regressed anchors.

To some extent, our proposed OSCAR can be viewed as a continuation of this concept, but it differs in two ways:
% (1) First, OSCAR is a fully convolutional anchor-free detector with few hyper-parameters related to anchor boxes;
(1) OSCAR abandons the commonly used divide-and-conquer strategy, which assigns high-level feature maps to large objects and low-level feature maps to small objects, and instead treats multi-scale predictions as progressive approximations of the target.
(2) to the best of our knowledge, this is the first time that top-down cascade bounding box refinement has been applied.
% These differences make OSCAR a unique and effective method for detecting small targets in infrared images.

% Recently, some works attempt to solve object detection in a per-pixel prediction fashion, analog to semantic segmentation.
% By eliminating the predefined set of anchor boxes, such methods, e.g., FCOS \cite{ICCV19FCOS}, FovealBox \cite{TIP20FoveaBox}, and CenterNet \cite{arXiv19CenterNetKeypoint}, completely avoid the complicated computation related to anchor boxes such as calculating overlapping during training as well as all the corresponding hyper-parameters.
% A trend for these one-stage detectors is to introduce an individual prediction branch to estimate the quality of localization such as center-ness \cite{ICCV19FCOS} or IoU \cite{IVC20IoUAwareDet,NIPS20GFLv1,CVPR21GFLv2}, where the predicted quality facilitates the classification to improve detection performance.
% While these approaches and the proposed OSCAR are both anchor-free detectors, our OSCAR differs in the following two ways:
% (1) in this paper, we revisit the current anchor-free label assignment scheme and highlight the potential for mislabeling targets as background samples. To this end, we design a label assignment scheme based on pseudo-boxes that is decoupled from bounding box targets.
% (2) the proposed OSCAR performs the classification and regression procedure iteratively in an effort to get higher localization performance.

% !TEX root = ../main.tex

% \section{DeepInfrared Eco-System: Fundamental Building Blocks}
\section{DeepInfrared Eco-System}
\label{sec:ecosystem}

% In this section, we introduce our DeepInfrared eco-system, which provides a complete benchmark for research on infrared small target detection, including a high-quality dataset, a reliable evaluation metric, and a toolkit for implementing and evaluating algorithms.

% we introduce our DeepInfrared eco-system, which aims to become a complete benchmark for the research of infrared small target detection.
% It consists of (1) SIRST-V2, a new real-world dataset, (2) normalized contrast, a new metric to evaluate the locating accuracy, and (3) DeepInfrared, an open-source toolkit created for this task.

\subsection{SIRST-V2: Pay More Attention to Urban Scenarios}

\begin{figure*}[htbp]
  \centering
  \includegraphics[width=0.975\textwidth]{"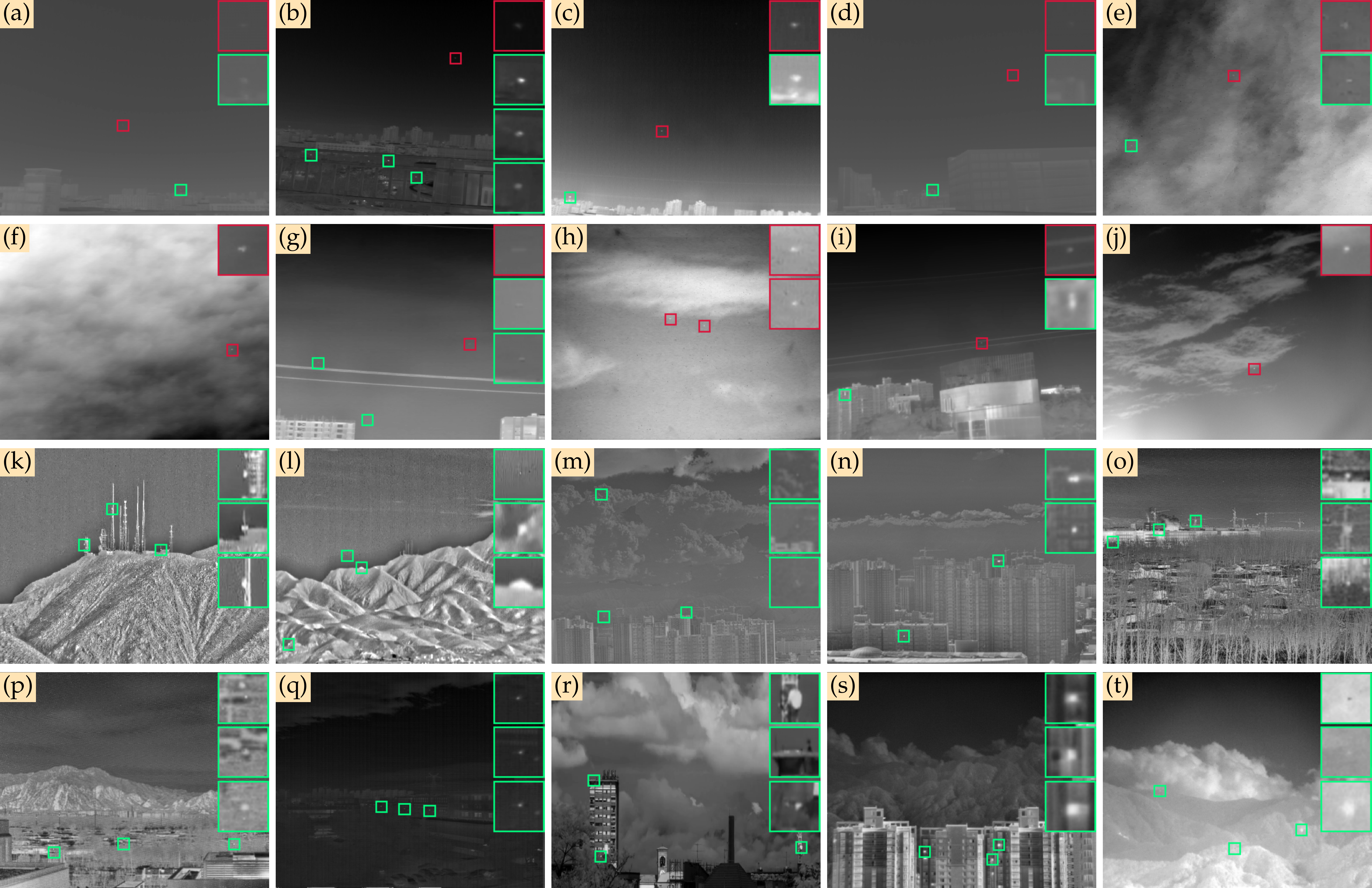"}
  \caption{The representative infrared images from the SIRST-V2 dataset with various backgrounds, excluding many simple cases.
  The {\color{WildStrawberry} \textbf{red boxes}} indicate {\color{WildStrawberry} \textbf{real targets}}, while the {\color{YellowGreen} \textbf{green boxes}} indicate {\color{YellowGreen} \textbf{background distractors}} that are similar to the targets.
  Compared to SIRSTv1, SIRSTv2 includes more urban scenes with many cranes, streetlights, and other non-target background interference, which require a high-level semantic understanding to distinguish.
  % Compared to SIRSTv1, SIRSTv2 includes more urban scenes with many cranes, streetlights, and other non-target background interference.
  % The presence of many target-like inferences in the background requires a high-level semantic understanding of the entire image in order to distinguish them.
  % For better visualization, the demarcated area is enlarged and can be better seen by zooming on a computer screen.
  }
  \label{fig:dataset}
  \vspace{-1\baselineskip}
\end{figure*}

First, we introduce our new SIRST-V2 dataset, which is the largest highest-resolution dataset for single-frame infrared small target detection that we are aware of. It is a milestone update of our previous SIRST-V1 dataset \cite{WACV21ACM} and contains 1024 representative images, mostly at a resolution of $1280 \times 1024$. These images have been extracted from real-world videos of various scenarios and provide a challenging testbed for infrared small target detection in complex real-world scenarios.
% Our dataset can be used to train and evaluate the performance of detection algorithms in complex real-world scenarios.

\cref{fig:dataset} illustrates some representative images from our new SIRST-V2 dataset.
In these images, it is evident that most infrared small targets are dim and difficult to distinguish from their backgrounds due to the presence of clutter. Additionally, the newly added urban scenes in SIRST-V2 contain many background interference elements such as cranes and street lights that are similar in appearance to the targets. This makes it challenging for traditional approaches that rely on target saliency or low-rank and sparse decomposition to effectively identify the targets in these complex scenarios. A more effective approach would require the use of a network with a high-level semantic understanding of the entire image to accurately distinguish targets from non-target interferences.
\vspace{-.5\baselineskip}

\subsection{Normalized Contrast: A Measure for Locating Accuracy}

\begin{figure*}[htbp]
  \centering
  \subfloat[Sensitivity analysis of IoU]{
    \includegraphics[width=.32\textwidth]{
      "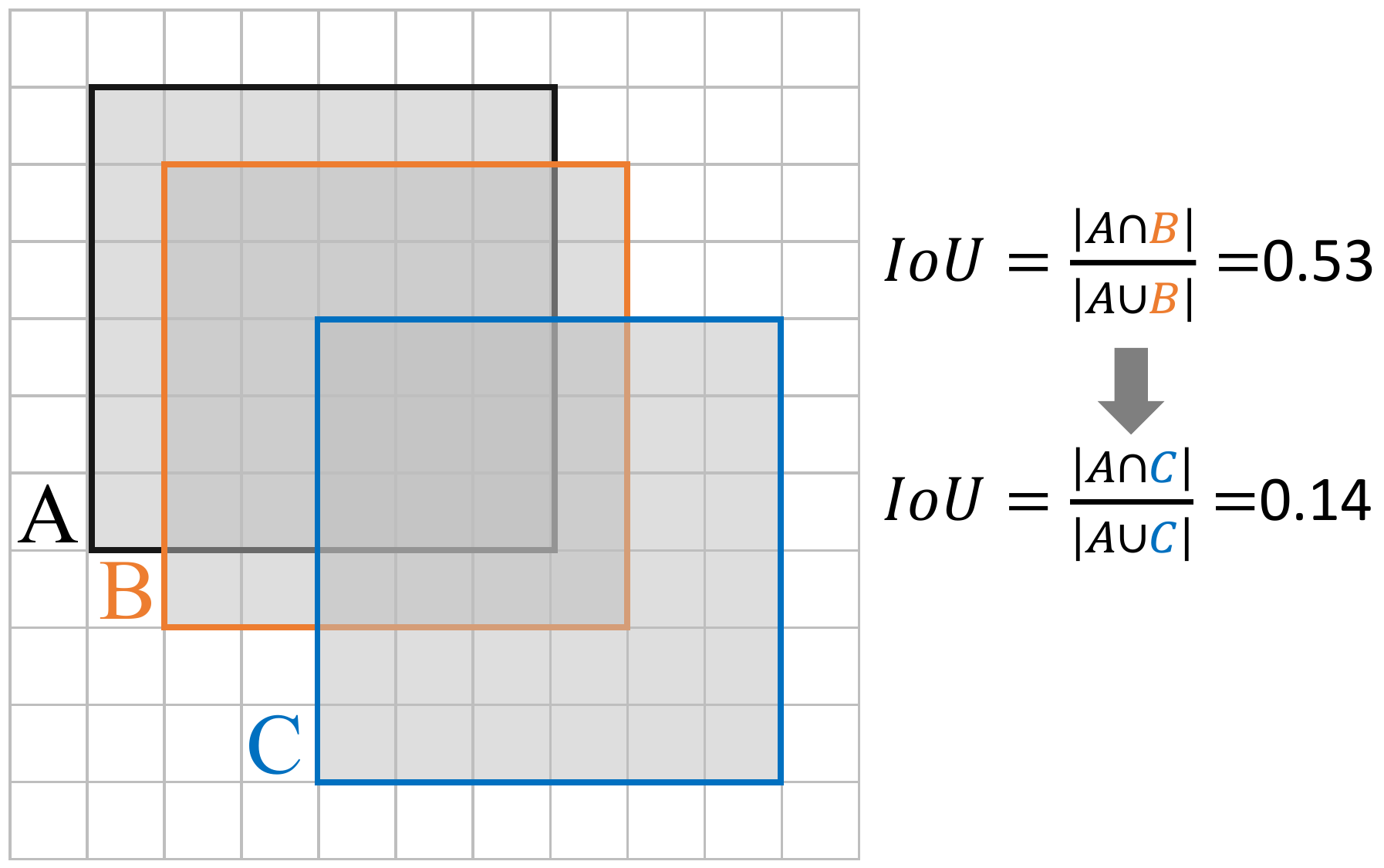"}
  }
  \subfloat[Dispersion effect and labeling ambiguity]{
    \includegraphics[width=.64\textwidth]{
      "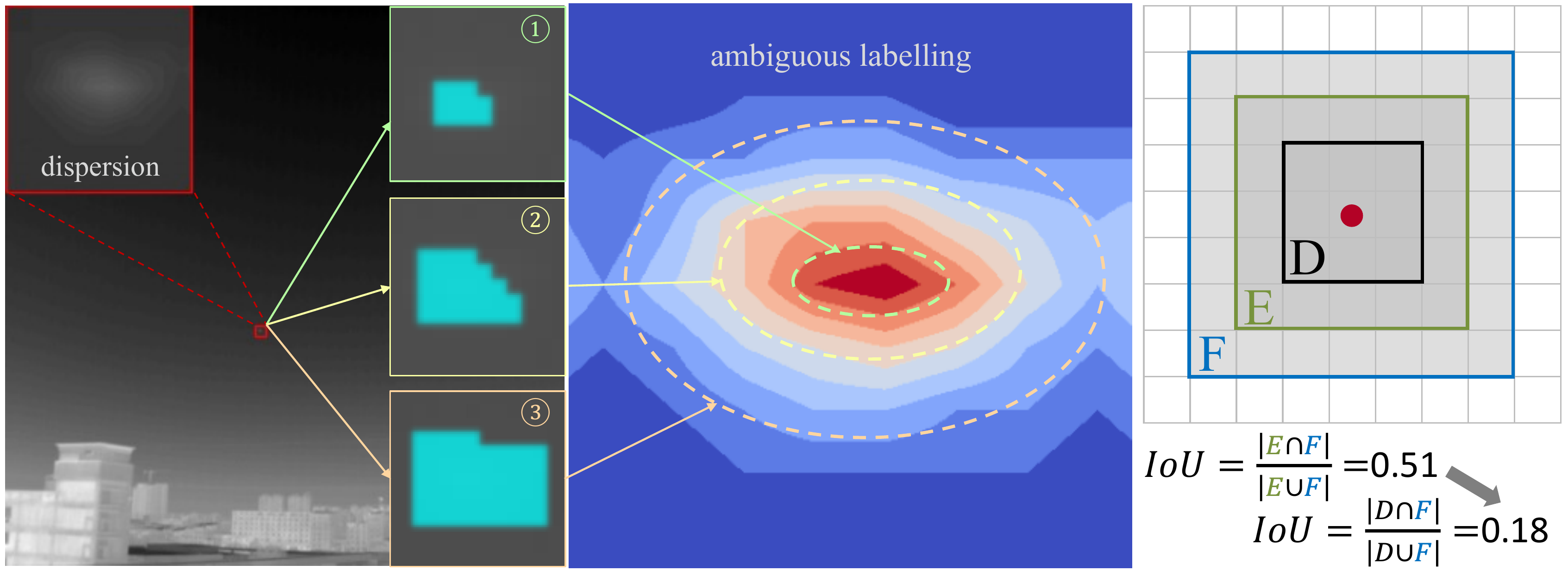"}
  }

  \caption{
    Drawbacks of IoU metric for infrared small targets. (a) The sensitivity analysis of IoU. Each grid denotes a pixel, box $A$ indicates the groundtruth box, box $B$ and $C$ stand for the predicted box with 1 pixel and 3 pixels diagonal deviation, respectively. (b) Illustration of the dispersion effect and labeling ambiguity.
    Because the boundaries of infrared small targets cannot be precisely defined due to long-range imaging, the groundtruth labels (blue masks) are ambiguous and can be annotated in a variety of ways, as illustrated in the figure (\textcircled{\raisebox{-0.7pt}{1}}, \textcircled{\raisebox{-0.7pt}{2}}, \textcircled{\raisebox{-0.7pt}{3}} three cases).
    Box $D$, $E$ denote the possible groundtruth boxes, box $F$ indicates the predicted box.
    It can be seen that a minor location deviation in either predicted box or groundtruth box will lead to notable IoU drop, resulting in inaccurate localization evaluation.
    Despite their different IoU values, box $E$ and $F$ share the same target centroid.}
  \label{fig:ambiguity}
  \vspace{-1\baselineskip}
\end{figure*}

% The IoU metric is highly skewed to large objects, very sensitive for evaluating the performance of infrared small targets due to their tiny size, e.g., $3 \times 3 \sim 12 \times 12$ (see \cref{fig:ambiguity} (a)).
The IoU metric is highly skewed to large objects, very sensitive for infrared small targets due to their tiny size (see \cref{fig:ambiguity} (a)).
A small perturbation in either predicted box or groundtruth box can cause a significant change in the IoU value, which leads to unreliable results.
However, in real-world applications, the main objective is to accurately locate the centroids of the infrared small targets so that they can be used for tracking. Therefore, IoU is only an intermediate representation and does not accurately reflect the overall detection performance. As such, it is necessary to design a new evaluation metric specifically for this task.

\begin{figure*}[htbp]
  \centering
  \includegraphics[width=.98\textwidth]{"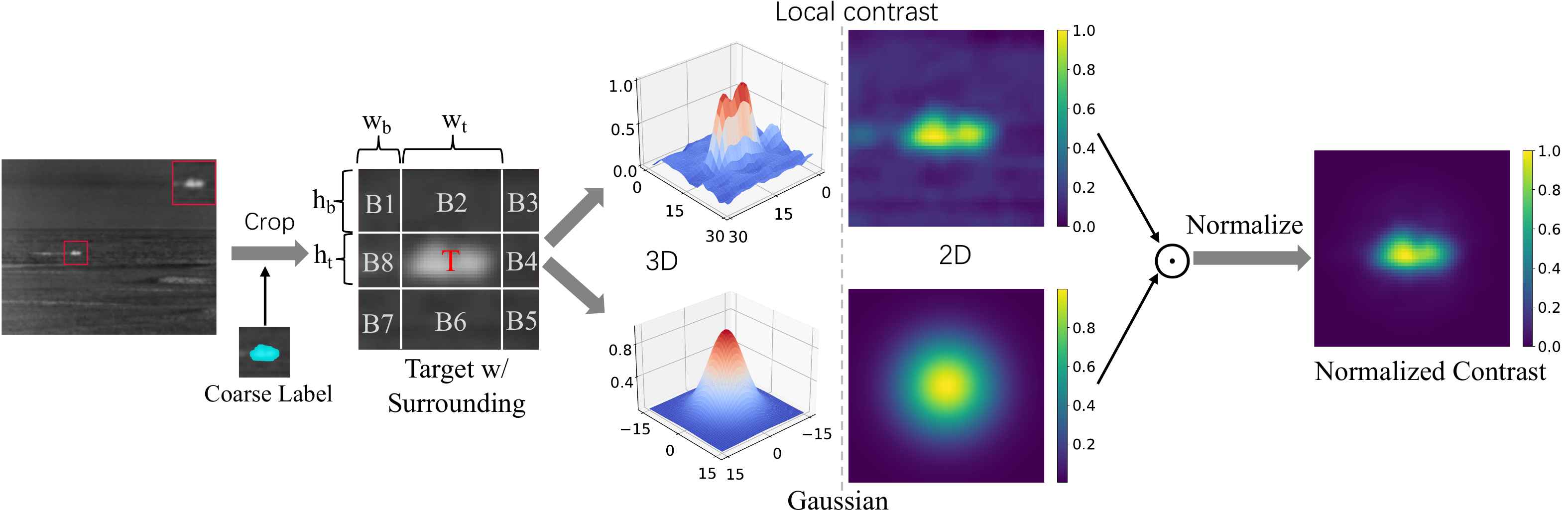"}
  \caption{
  The generation of normalized contrast (NoCo) map.
  First, the linear local contrast is calculated to provide a distribution that matches the appearance of the target. Then, a Gaussian is applied to give preference to the geometric center of the target. Finally, a coarse label is used to semantically normalize the labeled target region and some background pixels, and the rest of the values are set to 0.
  This process results in a NoCo map that is not sensitive to disturbances in the bounding box, making it a reliable representation of the target.
  % This is important for infrared small targets because there is no such a bounding box that can be perfectly aligned with the target due to the dispersion effect by long-range imaging.
  }
  \label{fig:noco-generation}
  \vspace{-1\baselineskip}
\end{figure*}

As far as we are concerned, a good localization quality measure for infrared small targets should have two key properties. First, it should be robust to label ambiguity, meaning that it can accurately reflect the location of the target even when the labels are not perfectly aligned with the target. Second, it should be able to directly reflect the accuracy of the target centroid, meaning that it should provide a clear indication of the precision of the target's location.

In our opinion, if the issues of computational efficiency and semantic normalization are addressed, the local contrast map has the potential to be a good quality measure for infrared small targets, as it offers a soft distribution that is adaptable to the target's shape and appearance.
Despite the numerous definitions of local contrast, their computational efficiency is very low, as can be seen in \cref{tab:sota} for local contrast measure (LCM) \cite{TGRS13LCM}, multi-scale patch-based contrast measure (MPCM) \cite{PR16MPCM} and weighted local difference measure (WLDM) \cite{TGRS16WLDM} in terms of frame per second (FPS). This is because these methods do not have groundtruth information, and therefore require complex nonlinear filtering operations to suppress background noise.
In fact, due to the availability of bounding box information, using only the linear normalization of the target bounding box region can also generate high-quality local contrast maps.
As a result, we propose to construct a new representation, called \emph{normalized contrast}, by normalizing the local contrast map according to the bounding box annotation, as shown in \cref{fig:noco-generation}.
Since the groundtruth box is known, we only need to calculate the normalized contrast for the bounding box area and its surroundings, and set the rest of the values to 0.
Given a groundtruth box $T = (c_x, c_y, h_t, w_t)$, we can obtain a larger region $R = (c_x, c_y, h_t+2h_b, w_t+2w_b)$ by extending a specific number of pixels in all directions ($h_b$ in Y-axis, $w_b$ in X-axis).
$\mathcal{C}_i = R_i - \min(R)$ is the unnormalized local contrast. $R_i$ denotes the grayscale of the $i$-th pixel in the region $R$ and $\min(R)$ indicates the minimum grayscale in region $R$.
Here we set $h_t = \gamma h_b, w_t = \gamma w_b$, where $\gamma \in (0, 1]$ is a constant.
The normalized contrast is defined as

\begin{equation}
  % \mathrm{NoCo}
  n_i = \frac{C_i \odot G_\sigma- \min(C_i \odot G_\sigma)}{\max(C_i \odot G_\sigma) - \min(C_i \odot G_\sigma)}
\end{equation}
The Gaussian distribution, denoted by $G_\sigma$, is used to impose a central preference on the map, which helps to suppress the background clutters in the surrounding region.
% . This is because the pixel with the highest local contrast for an infrared small target may not necessarily be located at the geometric center of the target.
The normalized contrast map depicts the normalized distance from a given location to the center of the infrared small target, providing a clear indication of the precision of the target's location.
In other words, the groundtruth normalized contrast map is actually a look-up table. Given a center point of a predicted bounding box, we can obtain a normalized contrast value for evaluating the localization accuracy of the predicted bounding box.

Note that our normalized contrast measure differs from the center-ness measure used in the FCOS in several key ways.
\begin{enumerate}
  \item First, the centroid of infrared small targets is typically annotated with the pixel of the highest signal-to-clutter ratio, rather than the geometric center of bounding boxes.
  The requirement for centroid locating accuracy is often at the sub-pixel level (smaller than 1 pixel), and using the center of the bounding box as the centroid may result in an error of several pixels.
  To reflect this domain knowledge, the normalized contrast is designed to directly reflect the signal-to-clutter ratio of each pixel, which aligns with the definition of the centroid.

  \item Second, the center-ness is purely a measure of coordinate distances, and a small perturbation of the bounding box can cause a significant disturbance.
  In contrast, our normalized contrast is a soft distribution adaptive to the appearance of infrared small targets.
  By replacing the delta distribution (step function) of the bounding box with a slowly-changing soft distribution, we are able to provide a more stable measure of target localization.

  % \item Third, the calculation of center-ness requires that the feature point fall within the ground-truth bounding box. However, for infrared small targets, this may not always be the case. As shown in \cref{fig:asap}, when the target is smaller than the stride, the center-ness does not exist. In contrast, our normalized contrast is calculated on the original input image, which is essentially a softened segmentation mask. This means that the normalized contrast is always available, regardless of how the feature map is downsampled.
\end{enumerate}
% Overall, these differences make our normalized contrast measure a more robust and reliable tool for localization of infrared small targets.

The mean Normalized Contrast Average Precision (mNoCoAP) is a metric that we propose to fairly and precisely assess the performance of algorithms for detecting infrared small targets. It is similar to the mean Average Precision (mAP) metric, which is widely used in generic object detection.
The mNoCoAP is calculated based on the Average Precision, which is the area under the Precision-Recall curve. The key difference between the mAP and our mNoCoAP lies in the definition of the True Positive criterion. In the mNoCoAP, we replace the IoU with the NoCo value of the predicted bounding box. This means that a prediction will be classified as a True Positive if the predicted NoCo value is above a certain threshold.
Following the common practice in the COCO dataset, the mNoCoAP is an average of nine levels of $\delta$, which controls the desired centroid locating precision, ranging from 0.1 to 0.9 with a step size of 0.1
% These levels range from 0.1 to 0.9, with a step size of 0.1.
% By using the mNoCoAP metric, we are able to accurately and fairly evaluate the performance of algorithms for detecting infrared small targets.

We have chosen the mNoCoAP as the main evaluation metric for our DeepInfrared benchmark because it offers several advantages over existing metrics.
\begin{enumerate}
  \item First, the mNoCoAP is a uniform evaluation metric that supports fair and preference-free evaluation of the performance of different detection paradigms, such as bounding box regression, semantic segmentation, and background suppression.
  % This means that it can be used to compare the performance of different algorithms without bias towards any particular approach.

  \item Second, the mNoCoAP is different from the receiver operating characteristic (ROC) curve in that it can not only be used to evaluate the performance of producing as few false positives or false negatives as possible, but it can also be used to evaluate the centroid locating accuracy of the tested method. This is an important consideration when dealing with infrared small targets, as the precision of the target's location is often critical for successful detection.

  \item Third, the mNoCoAP is more resistant to label ambiguity than other metrics because it softens the harsh truncated boundary of an ambiguous ground-truth bounding box. This is especially important in the case of long-range imaging, where label ambiguity can be a significant issue.
\end{enumerate}
% Overall, the use of the mNoCoAP metric offers a number of advantages for evaluating the performance of algorithms for detecting infrared small targets.

\subsection{DeepInfrared Toolkit: An Open-Source Benchmark}

In the field of generic computer vision, there are many open-source object detection toolboxes, such as MMDetection and GluonCV, that implement and integrate state-of-the-art detection algorithms. These toolboxes are widely used by researchers to reproduce experiments and improve algorithms.
However, in the case of infrared small targets, there is currently no open-source toolbox available. This lack of a common platform makes it difficult for researchers to reproduce experiments and compare different algorithms, which can hinder progress in this field.

To this end, we have developed DeepInfrared, a deep learning toolkit specially developed for the infrared small target detection task based on PyTorch. Compared with its counterparts in generic computer vision, our DeepInfrared toolkit differs in the following two points:
\begin{enumerate}
  \item DeepInfrared provides well-trained models, training scripts, and logs of state-of-the-art infrared small target detection algorithms. This allows researchers to easily reproduce experiments and compare different algorithms.
  \item DeepInfrared offers more flexible choices for backbones and necks, and has implemented various dataset loaders, attention blocks, and data augmentation pipelines. It also includes evaluation metrics specifically designed for infrared small target detection, and supports multiple annotation formats, such as semantic segmentation, bounding boxes, and normalized contrast.
\end{enumerate}

% We hope our DeepInfrared can be a valuable resource for researchers working on infrared small target detection and help to accelerate progress in this field.

% !TEX root = ../main.tex
% \bibliography{../../../ref-bib/all-refs.bib}

\section{One-Stage Cascade Refinement Networks}

% In this section, we propose our OSCAR network to improve the detection of infrared small targets.
% First, we design the ASPB label assignment scheme to avoid the issue of small targets being missed.
% Next, we propose the Top-Down Cascade Refinement Head, which reformulates the multi-level predictions of a feature pyramid as approximations of cascaded soft region proposals in a coarse-to-fine order towards the same target.
% Finally, we incorporate a semantically normalized contrast branch, which helps suppress low-quality detected bounding boxes introduced by pseudo-boxes and significantly improves overall performance.

\subsection{All Scale Label Assignment with Pseudo-Boxes}

% anchor-based 和 anchor-free 的选择
% Depending on how the positive and negative training samples are defined, one-stage detector can be divided into anchor-based and anchor-free methods.
% Anchor-based detectors calculate the intersection-over-union (IoU) scores between a set of pre-defined dense anchor boxes and ground-truth bounding boxes to assign labels to the training samples.
% The detection performance is sensitive to hyper-parameters related to anchor boxes, such as their numbers, aspect ratios, sizes, and the IoU threshold.
% Especially, for infrared small targets, one or two pixel shifts in the anchor boxes can significantly influence the IoU scores and further degrade the final detection performance, due to the small sizes of ground-truth bounding boxes.
% On the contrary, anchor-free methods regards the center (e.g., the center point or part) of object as foreground to define positives, which avoid all hyper-parameters related to anchor boxes.

Let $F_{i} \in \mathbb{R}^{H \times W \times C}$ be the feature maps at layer $i$ of a feature pyramid and $s_i$ be the total down-sampling stride of this layer.
A ground-truth bounding box is defined as $B_\mathrm{g}=\left(c_x, c_y, h, w\right)$, where $\left(c_x, c_y\right)$ denotes the center of the bounding box, and $h, w$ are its height and width.
To avoid using hyper-parameters related to anchor boxes, we choose a one-stage anchor-free method as the baseline, which generally regards the center (e.g., the center point or part) of the object as foreground to define positives \cite{TPAMI22GFL}.
However, such center-based label assignment has an implicit assumption that the positive area inside the bounding box of size $h \times w$ must be larger than the square of the stride $s_i$, namely, $h \times w \ge s_i^2$, so that at least one feature point would be labeled as positive.

\begin{figure*}[htbp]
  \centering
  \includegraphics[width=.85\textwidth]{"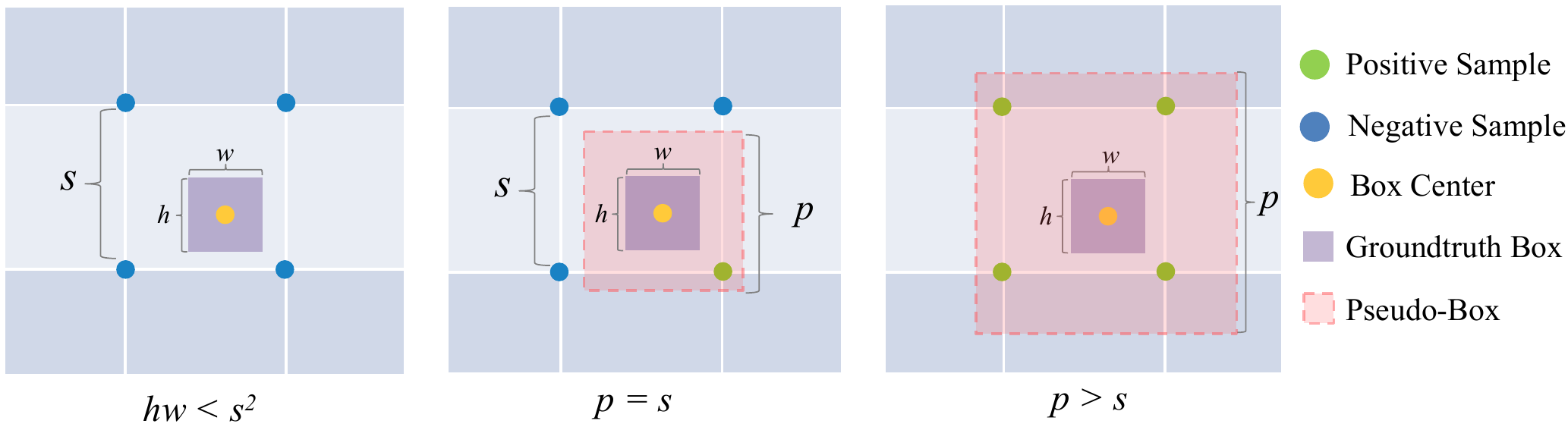"}
  \caption{
    Illustration of pseudo-box based label assignment.
    \textbf{left:} The ground-truth box based label assignment defines positive samples for each output spacial position by checking whether it falls into the box.
    \textbf{middle:} In the pseudo-box based label assignment, each real target is assigned to at least one feature point by using a pseudo-box that is larger than the ground-truth box.
    \textbf{right:} The class imbalance problem between positive and negative samples will be resolved by increasing the pseudo-box size, but at the expense of increasing the number of low-quality predicted boxes.}
  \label{fig:asap}
  \vspace{-1\baselineskip}
\end{figure*}

This is a valid assumption for generic object detection, but not for infrared small targets, whose sizes generally vary from $2\times 2$ to $12 \times 12$.
For example, in FCOS, the stride $s$ of the finest layer is $8$. If $hw < s^2$ for a given target, the bounding box of that target may not cover any feature points, leading to it being mislabeled and resulting in low detection performance, as shown in \cref{fig:asap}. This is because a large portion of infrared small objects in a certain scale are not being trained.

% rethink
This motivates us to \emph{re-examine the label assignment for infrared small targets}.
In fact, label assignment can be further divided into scale assignment and spatial assignment.
% 通用视觉怎么操作的
In generic object detection, the scale assignment assigns targets of different sizes to the appropriate feature pyramid level for detection, and the spatial assignment then selects positive sample locations at the selected level (e.g., center sampling in FCOS \cite{ICCV19FCOS}).
% 红外的特点
However, infrared small targets do not exhibit large scale variation. Instead, many targets are missed if the stride of the prediction layer is larger than the target.
%  This is a notable difference from generic object detection.

Motivated by these observations, we propose an All-Scale Pseudo-Box (ASPB) based scheme to customize the label assignment for anchor-free infrared small target detection to improve detection performance.
The main idea is to decouple the ground-truth target size from the label assignment.
Specifically, for scale assignment, we no longer consider the small targets predicted by each scale as competitors, but rather as coarse-to-fine approximations under different computational budgets.
To this end, we relax the scale constraints on the label assignment and treat all target boxes at all scales as positive samples.
For the spatial assignment, we introduce a pseudo-box with the same center and size of $p \times p$ ($p \ge s$) to replace the ground-truth bounding boxes of the tiny targets whose sizes are smaller than the feature map stride, \textit{i.e.}, $h \times w < s^2$.
This allows us to assign positive labels to more feature points, improving detection performance for infrared small targets.

Given a ground-truth bounding box $(c_x, c_y, h, w)$, the pseudo-box $B_\mathrm{p}$ is defined as
\begin{equation}
  B_\mathrm{p}=\left\{\begin{array}{lc}
    (c_x, c_y, h, w), & \text { if } hw > p^2 \\
    (c_x, c_y, p, p), & \text { otherwise }
  \end{array}\right.
  \end{equation}
As illustrated in \cref{fig:asap}, when $p = s$, at least one feature point will be labeled as positive and the target mislabeling issue is resolved. As $p$ increases, \textit{i.e.}, $p > s$, more neighboring feature points will be labeled as positive samples, which further relieves the sample imbalance issue but at the cost of introducing more low-quality detected bounding boxes.

\subsection{Top-Down Cascade Refinement Head}

The performance of a one-stage detector is also affected by the severe sample imbalance and inaccuracy of bounding box regression. The tricky problem is that the solutions to these two issues have opposite requirements in terms of scale.
With our ASPB label assignment scheme, making predictions on the high-level feature maps of the feature pyramid is a straightforward way to solve the class imbalance problem because the number of negative samples will drop exponentially as the down-sampling stride increases. However, this solution is not ideal for improving bounding box regression accuracy because high-level feature maps have low spatial resolution and are not suitable for precise localization.
Instead, because it is simpler to regress a local box than a remote one, the bounding box regression task favors low-level fine feature maps over high-level coarse feature maps.
Simply dividing the classification and localization tasks into different scales, namely classifying the feature points at the high level and regressing the bounding box at the low level, is not a workable solution because the one-to-many mapping relationship between multi-scale feature points can lead to high rankings for low-quality bounding boxes, degrading the final detection performance.

\begin{figure*}[htbp]
  \centering
  \includegraphics[width=.95\textwidth]{"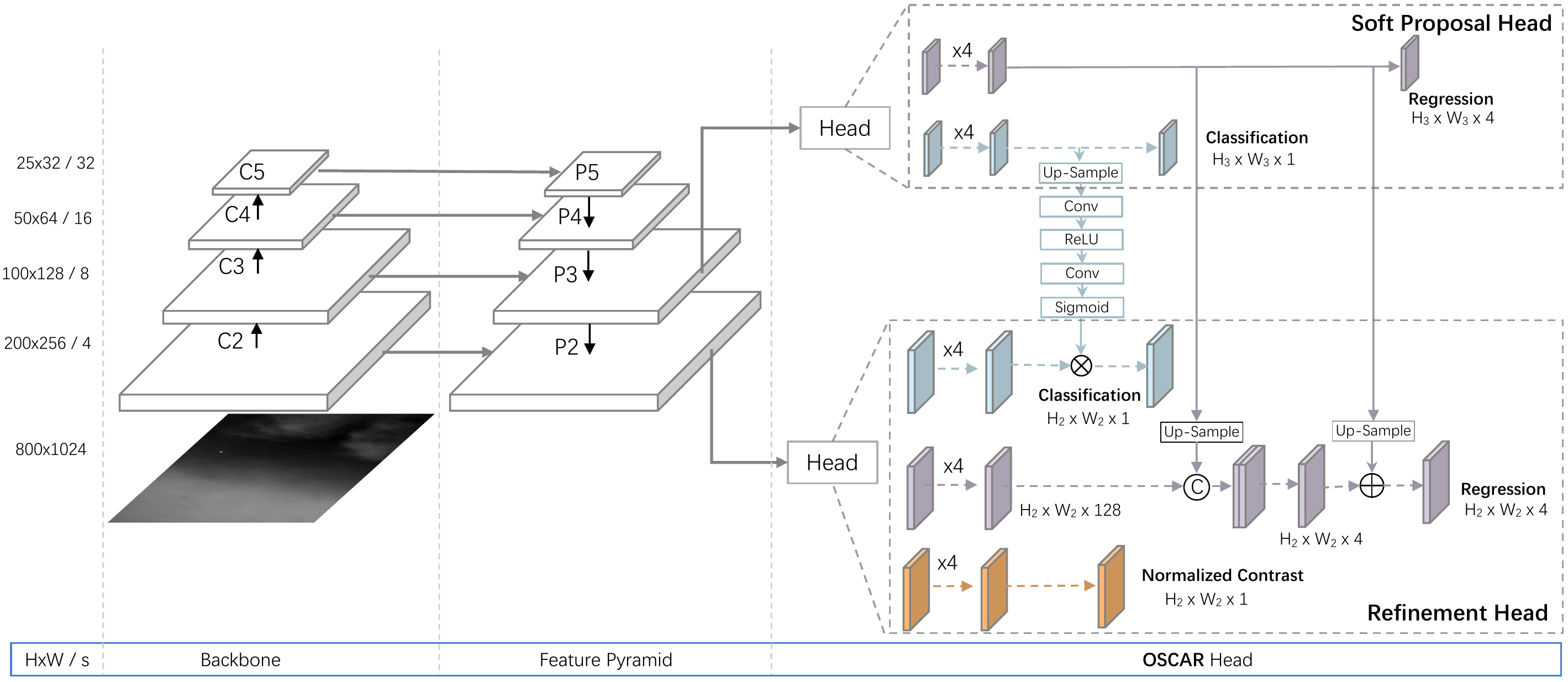"}
  \caption{
  The network architecture of OSCAR. In OSCAR, a feature pyramid's multi-level predictions are reformulated as approximations of cascaded soft region proposals in a coarse-to-fine order towards the same infrared small target. In addition, a normalized contrast branch is augmented to suppress low-quality predicted bounding boxes caused by pseudo-boxes.
  % C2, C3, C4, and C5 denote the feature maps of the backbone network and P2 to P5 are the feature levels used for the final prediction.
  $H \times W$ is the height and width of feature maps. "/s" ($s = 4, 8, 16, 32$) is the down-sampling ratio of the feature maps at the level to the input image. All the numbers, for instance, are computed using an input of $800 \times1024$.}
  \label{fig:oscar-arch}
  \vspace{-1\baselineskip}
\end{figure*}

The proposed OSCAR Head is a solution to the issue of combining the efficiency of a one-stage detector with the merits of region proposal and cascaded bounding box refinement from two-stage detectors. The OSCAR Head consists of a high-level soft proposal head and a low-level refinement head, as shown in \cref{fig:oscar-arch}.
The soft proposal head predicts a classification score $\boldsymbol{p}^{h}{x, y}$ and a coarse bounding box $\boldsymbol{b}^{h}{x, y}$ for each feature point $(x,y)$. These coarse boxes are then passed as initial guesses with top-down modulation scores to the refinement head, where they are further refined by predicting a category score $\boldsymbol{p}^{l}{x, y}$ and a refined bounding box $\boldsymbol{b}^{l}{x, y}$.

\subsubsection{Top-Down Score Modulation}
To address the extreme imbalance between small targets and background during training, we propose a top-down score modulation mechanism as a soft region proposal module.
This module is designed to improve the accuracy of the detector by using the more accurate score of the coarse layer to modulate the less accurate score of the fine layer.
Compared to the serialized region proposal operation in two-stage detectors, which involves cropping feature maps and performing ROI Pooling, our soft region proposal can be implemented in a single forward pass and is therefore more efficient. This allows for parallel processing, which can improve the overall performance of the detector.

Given a coarse feature map $X^h$ of size $(H^h, W^h)$ and a fine feature map $X^l$ of size $(h^l, w^l)$, both are the last feature maps before prediction in corresponding layers.
The top-down score modulation weight is defined as follows:

\vspace{-1\baselineskip}
\begin{equation}
  T(X^h) = \sigma(\mathrm{Conv}_2(\delta(\mathrm{Conv}_1(\mathrm{Up}(X^h)))))
\end{equation}
where $\mathrm{Conv}_1$ and $\mathrm{Conv}_2$ are convolutional layers whose kernel sizes are $\frac{C}{4} \times C \times 3 \times 3$ and $C \times \frac{C}{4} \times 3 \times 3$, respectively.
The $\delta$ denotes the Rectified Linear Unit (ReLU) and $\sigma$ is a Sigmoid function.
$\mathrm{Up}(\cdot)$ denotes the up-sampling operation to interpolate the high-level feature maps to the same size of low-level feature maps.
Then the modulated low-level classification scores can be obtained via
\begin{equation}
  S^l = \sigma(\mathrm{Conv}(T(X^h) \odot X^l))
\end{equation}
where $\odot$ is the element-wise multiplication.
% Here, the semantic information provided by high-level features is utilized to filter out well-classified negative samples in a soft manner.
By modulating the scores of the fine layer with the more accurate scores from the high-level features, the top-down modulation approach effectively eliminates false negatives and reduces the impact of class imbalance on the detector's performance.
% This allows the detector to focus on accurately identifying infrared small targets, resulting in improved performance.

\subsubsection{Top-Down Cascaded Regression}

One-stage methods typically rely on a single step of regression using different feature layers to predict the locations and sizes of objects with varying scales. However, this approach can be ineffective in certain challenging scenarios, particularly for infrared small targets. To address this limitation, we propose a cascaded regression strategy for more accurate prediction of target locations and sizes.
To improve the regression accuracy in the fine-level feature maps, we use the high-level layer to first predict coarse locations and sizes of underlying infrared small targets. This initial coarse prediction serves as a learnable anchor for the fine-level layer to further refine the predictions.

The decoded coarse bounding box is denoted as $B^{\mathrm{h}} = \left(x_0^{h}, y_0^{h}, x_1^{h}, y_1^{h}\right)$. To align with the next-level refined feature maps, we first need to up-sample the coarse bounding boxes. Here, we denote the interpolated bounding box as $\tilde{B}^{\mathrm{h}} = \left(\tilde{x}_0^{h}, \tilde{y}_0^{h}, \tilde{x}_1^{h}, \tilde{y}_1^{h}\right)$.
To further produce precise object locations and sizes, we feed both the coarse features and the coarse boxes to the refinement head.
Formally, if a feature point is associated to a groundtruth box $B_\mathrm{g}=\left(x_0, y_0, x_1, y_1\right)$, the regression targets for this feature point can be formulated as
% \vspace{-.25\baselineskip}
\begin{equation}
  \begin{aligned}
  \Delta x_0 & =(x_0-\tilde{x}_0^{h})/s, \quad \Delta x_1=(x_1-\tilde{x}_1^{h})/s \\
  \Delta y_0 & =(y_0-\tilde{y}_0^{h})/s, \quad \Delta y_1=(y_1-\tilde{y}_1^{h})/s.
  \end{aligned}
\end{equation}
where $s$ is the stride of the low-level feature map.
During inference, the regressed target $\hat{B}^{l}$ is predicted via concatenating the up-sampled feature maps, denoted as
\begin{equation}
  \hat{B}^{l} =  \mathrm{Conv}(\mathrm{Concat}(\mathrm{Up}(X^h), X^l))
\end{equation}
where $\mathrm{Concat}$ is the concatenating operation and $\mathrm{Up}$ denotes the feature map up-sampling.
To some extent, the coarse boxes can be viewed as dynamic anchors.

% Unlike RetinaNet, which directly uses regularly tiled anchors for detection, the proposed OSCAR head uses a two-step strategy, taking advantage of the complementary information provided by the high-level and low-level layers.
% This approach is particularly effective for infrared small targets, where the initial coarse estimates are crucial for the low-level layer to accurately refine the predictions.

\subsection{Normalized Contrast Prediction}

The proposed ASPB label assignment method significantly improves the class imbalance issue \cite{ICCV17FocalLoss}. However, it also introduces many low-quality predicted bounding boxes by assigning positive labels to feature points that are far away from the center of an infrared small target. To suppress these low-quality bounding boxes, we propose to add a new branch to the refinement head that predicts the localization quality score of a detected bounding box.
% This is similar to the "center-ness" branch in FCOS \cite{ICCV19FCOS} or the IoU branch in the IoU-Aware detector \cite{IVC20IoUAwareDet}.

It is important to note that the calculation of center-ness requires that the feature point be within the ground-truth bounding box. However, for infrared small targets, this requirement may not always be satisfied. As shown in \cref{fig:asap}, when the target size is smaller than the feature map stride, the center-ness cannot be calculated. In contrast, our normalized contrast is calculated using the original input image, which can be thought of as a softened semantic segmentation mask. No matter how the feature map is down-sampled, the normalized contrast can always be calculated.

% It is important to note that our normalized contrast differs from the center-ness in FCOS in the following two ways:
% \begin{enumerate}
%   \item The centroid of an infrared small target is typically annotated with the pixel of the highest signal-to-clutter ratio, rather than the geometric center of the bounding box. Therefore, the predicted bounding box with a higher local contrast should have a higher ranking during non-maximum suppression for this task, which is consistent with the mNoCoAP metric.
%   \item The calculation of center-ness requires that the feature point be within the ground-truth bounding box. However, for infrared small targets, this requirement may not always be satisfied. As shown in \cref{fig:asap}, when the target size is smaller than the feature map stride, the center-ness cannot be calculated. In contrast, our normalized contrast is calculated using the original input image, which can be thought of as a softened semantic segmentation mask. No matter how the feature map is down-sampled, the normalized contrast can always be calculated.
% \end{enumerate}

\textbf{Loss Function.}
% The loss function for OSCAR is composed of two parts: the loss in the soft proposal head and the loss in the refinement head. Both of these losses include a classification loss and a bounding box regression loss. In addition, the refinement head also has a localization quality loss, which is trained using the quality focal loss (QFL) loss \cite{TPAMI22GFL}.
% This is because the ground-truth normalized contrast has a range from 0 to 1. 
The overall loss is defined as follows:
% \vspace{-1\baselineskip}
\begin{equation}
  \begin{aligned}
  L &=\frac{1}{N^{h}_{\text {pos }}} \sum_{x,y}
    L_{\mathrm{cls}}\left(\boldsymbol{p}^{h}_{x,y}, c_{x,y}^{h,*}\right) +
    \mathbbm{1}_{\left\{c_{x,y}^{h,*}>0\right\}} L_{\mathrm{reg}}\left(\boldsymbol{t}^{h}_{x,y}, \boldsymbol{t}_{x,y}^{h,*}\right)\\
    & + \frac{1}{N^{l}_{\text {pos }}} \sum_{x,y}
    L_{\mathrm{cls}}\left(\boldsymbol{p}^{l}_{x,y}, c_{x,y}^{l,*}\right) +
    \mathbbm{1}_{\left\{c_{x,y}^{l,*}>0\right\}} L_{\mathrm{reg}}\left(\boldsymbol{t}^{l}_{x,y}, \boldsymbol{t}_{x,y}^{l,*}\right)\\
    & + \lambda L_{\mathrm{noco}}\left(n^{l}_{x,y}, n_{x,y}^{l,*}\right)
  \end{aligned}
\end{equation}
where $L_{\mathrm{cls}}$ is focal loss \cite{ICCV17FocalLoss}, $L_{\mathrm{reg}}$ the IOU loss, and $L_{\mathrm{noco}}$ is the quality focal loss as in \cite{TPAMI22GFL}.
$N_{\mathrm{pos}}$ denotes the number of positive samples and $\lambda$ being 1 in this paper is the balance weight for the regression loss.
$\mathbbm{1}_{\left\{c_{i}^{*}>0\right\}}$ is the indicator function, being 1 if the $c_{i}^{*}>0$ and 0 otherwise.

\textbf{Inference.} During testing, the predicted normalized contrast is combined with the two-level classification scores to calculate the final score for each detected bounding box. As a result, bounding boxes that are far from the center of an infrared small target may be given lower scores due to the effect of the normalized contrast. Therefore, the non-maximum suppression is likely to filter out these low-quality bounding boxes, which can greatly improve the detection performance.

% !TEX root = ../main.tex
% \bibliography{../../../ref-bib/all-refs.bib}

\section{Experiments}   \label{sec:experiment}

% To verify the effectiveness of the proposed OSCAR network, we conduct detailed ablation studies and compare it to state-of-the-art methods using the SIRST-V2 dataset and the mNoCoAP metric.

\subsection{Experimental Settings}

We have compared the OSCAR network to several model-driven methods for infrared small target detection. These methods include the method based on facet kernel and random walker (FKRW) \cite{TGRS19FKRW}, LCM \cite{TGRS13LCM}, MPCM \cite{PR16MPCM}, WLDM \cite{TGRS16WLDM}, infrared patch-image model (IPI) \cite{TIP13IPI}, non-negative IPI model via partial sum minimization of singular values (NIPPS) \cite{IPT17NIPPS}, reweighted infrared patch-tensor model (RIPT) \cite{JSTARS17RIPT}, and stable multi-subspace learning (SMSL) \cite{TGRS17SMSL}.
The detailed hyper-parameter settings for these methods are listed in \cref{tab:params}, which shows the parameters that were determined by an exhaustive search on the trainval set of the SIRSTv2 dataset.

\setlength{\tabcolsep}{4pt}
\begin{table*}[htbp]
\caption{Detailed hyper-parameter settings of model-driven methods for comparison.}
\label{tab:params}
\vspace*{-.5\baselineskip}
\centering
\small
\begin{tabular}{Sl Sl} 
\toprule
Methods & Hyper-parameter settings \\
\midrule
FKRW \cite{TGRS19FKRW} & $K=5$, $p=6$, $\beta=200$, window size: $11\times 11$ \\
LCM \cite{TGRS13LCM} & Cell Size: $3 \times 3$, threshold factor: $k=3$\\
MPCM \cite{PR16MPCM} & $N=1,3,...,9$, threshold factor: $k=13$\\
WLDM \cite{TGRS16WLDM} & $L = 4, m = 2, n = 2$, threshold factor: $k=10$\\
IPI \cite{TIP13IPI} & Patch size: 50$\times$50, stride: 20, $\lambda {\rm{ = }}L{\rm{/min(m,n}}{{\rm{)}}^{1/2}}$,$L=2.5$, $\varepsilon {\rm{ = 1}}{{\rm{0}}^{{\rm{ - 7}}}}$ \\
NIPPS \cite{IPT17NIPPS} &  Patch size: 50$\times$50, stride: 20, $\lambda = L/\sqrt{\min{(m,n)}}$, $L=1.2$, energy constraint ratio: $r=0.005$ \\
RIPT \cite{JSTARS17RIPT} &  Patch size: 50$\times$50, stride: 10, $\lambda = L/\sqrt{\min{(I,J,P)}}$, $L=0.001$, $h=10$, $\epsilon$=0.01, $\varepsilon = 10^{-7}$, threshold factor:$k=10$ \\
SMSL \cite{TGRS17SMSL} & Patch size: 30$\times$30, $\lambda = 2\times L/\sqrt{\min{(m,n)}}$, $L=5.0$, threshold factor: $k=1$ \\
\bottomrule
\end{tabular}
\vspace{-1\baselineskip}
\end{table*}

% $\lambda = L/\sqrt{\min{(m,n)}}$

We also have conducted a thorough comparison of the OSCAR network with several state-of-the-art methods for infrared small target detection. These methods include the Infrared Small-Target Detection U-Net (ISTDU) \cite{GRSL22ISTDUNet}, ACMNet \cite{WACV21ACM}, Attentional Local Contrast Networks (ALCNet) \cite{TGRS2021ALCNet}, FCOS \cite{ICCV19FCOS}, Faster R-CNN \cite{NIPS15FasterRCNN}, QueryDet \cite{CVPR22QueryDet}, and Gaussian Receptive Field based Label Assignment (RFLA) \cite{ECCV22RFLA}.
To ensure fair comparisons, we have used our DeepInfrared toolkit to produce all of the results. We have adopted the default hyper-parameters in all cases, and have used the standard 1x learning schedule (12 epochs) for the ablation study. The training and testing details for each method are based on the descriptions in the original papers.
% In other words, the backbone of ACMNet, ALCNet, and OSCAR is ResNet-18, while the backbone of FCOS, QueryDet, Faster R-CNN, and RFLA is ResNet-50.
All of the trained models, codes, and training logs, 
% including those for our OSCAR network and the compared methods, 
are available in our DeepInfrared toolkit. This allows researchers to easily reproduce our experiments and compare the performance of different algorithms.
% \vspace{-1\baselineskip}

% To facilitate reproducible research, we have included the separated target images produced by these methods in our DeepInfrared toolkit.

\subsection{Ablation Study}\label{subsec:ablation}

% To better understand the importance of each component in our proposed OSCAR detector, we have conducted a series of ablation studies.
% In these experiments, we evaluate a baseline method based on the ResNet-18 \cite{CVPR16ResNetV1} architecture, and then gradually add the ASPB label assignment, top-down cascade refinement, and normalized contrast branch to the model.

\subsubsection{Impact of Pseudo-Box Based Spatial Assignment}

One of the key performance bottlenecks of current anchor-free detectors for infrared small targets is the issue of target mislabeling caused by existing center-based label assignment methods \cite{ICCV19FCOS}. In this part, we show that this problem can be largely resolved by replacing the ground-truth target size with a pseudo-box in spatial label assignment.
To do this, we will investigate the following two sub-questions: (1) To what extent does the target mislabeling issue hurt the performance of a detector? (2)  What is the best pseudo-box size for infrared small target detection?
% \begin{enumerate}
%   \item 
%   \item
% \end{enumerate}

\begin{figure*}[htbp]
  \centering
  \subfloat[Anchor-free baseline]{
    \includegraphics[width=.45\textwidth]{
      "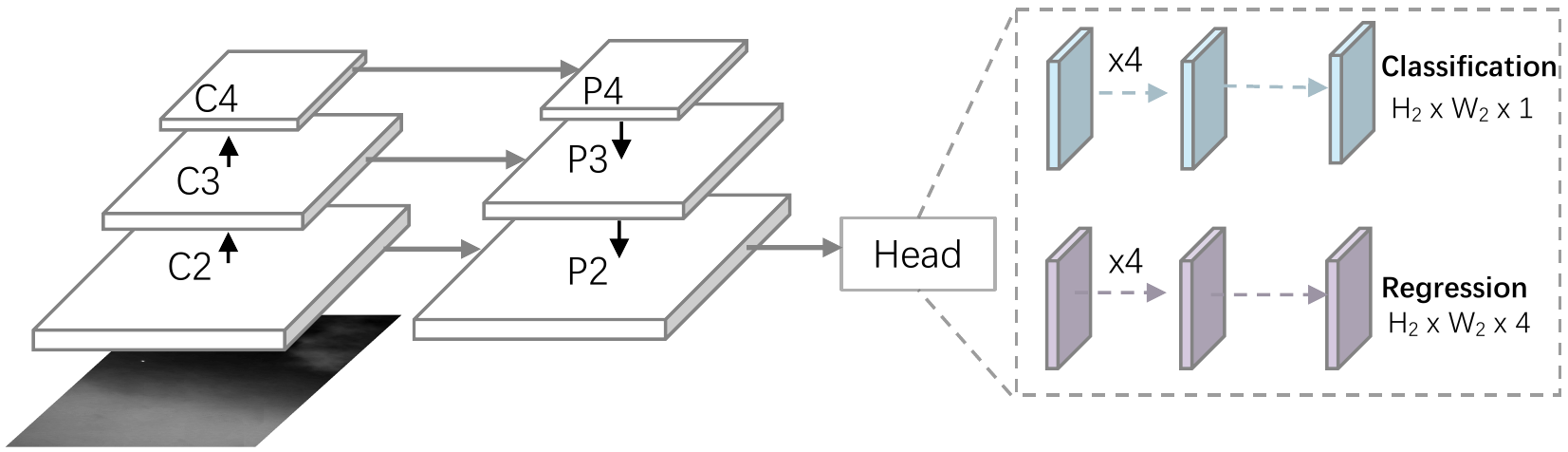"}
      \label{subfig:baseline-arch}
  }
  \subfloat[Decoupled head]{
    \includegraphics[width=.45\textwidth]{
      "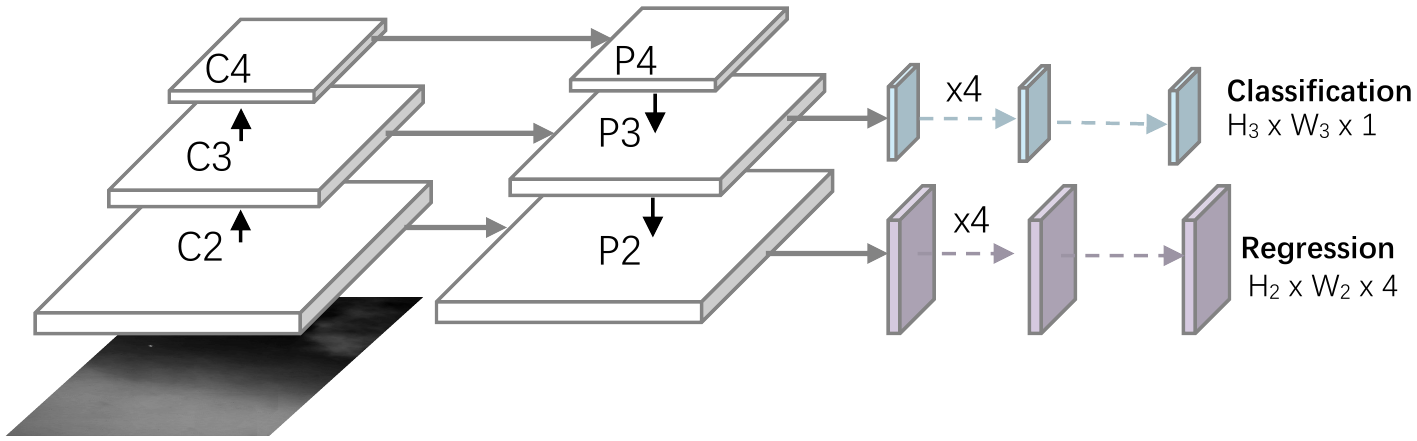"}
      \label{subfig:DecoupledHead}
  }
  \caption{The illustration of network architectures used for ablation study. (a) The anchor-free baseline detector, (b) Decoupled head that decouples the classification and regression tasks into separate levels. The high-level branch classifies feature points on coarse feature maps, while the low-level branch regresses bounding boxes on refined feature maps.}
  \label{fig:ablation-arch}
  \vspace{-1\baselineskip}
\end{figure*}

% To evaluate the impact of the target mislabeling issue, we used a simple anchor-free detector as the baseline for our ablation study. This baseline model can be seen as a simplified version of FCOS without the centerness branch and only one layer of prediction, as shown in \cref{fig:ablation-arch} (a).
We used a simple anchor-free detector as the baseline for this ablation study, which can be seen as a simplified version of FCOS without the centerness branch and only one layer of prediction, as shown in \cref{fig:ablation-arch} (a).
The results of our experiments are reported in \cref{tab:aspb}, where "None" denotes the use of the original center-based label assignment scheme, and $p=s$ means that at least one nearest neighbor feature point labeled as a positive sample.
As shown in the table, using a pseudo-box in spatial label assignment can improve the AP from 71.9\% to 75.6\%.
This gap indicates the performance degradation brought by the label noise caused by the center-based label assignment, which mislabels a portion of real targets as background samples.

% 代价是额外引入一个超参数
% 通过调节这个超参数，可以进一步提升性能
To determine the optimal pseudo-box size, we varied the value of $p$ in our experiments and measured the resulting detection performance. As shown in \cref{tab:aspb}, the detection performance increases as we increase the value of $p$ from $p=s$. This is because as more feature points are set as positive samples, the imbalance problem of positive and negative samples is relieved. However, when $p$ is too large, those positive feature points that are far from the target produce many low-quality detections, thus reducing the final detection performance. From the results, we consider $p=1.5s$ as a suitable size and use it in the following experiments.

\setlength{\tabcolsep}{4pt}
\begin{table}[htbp]
\centering
% \caption{Ablation Study for ASPB Label Assignment.}
\caption{Impact of pseudo-box sizes on our ASPB scheme.}
\label{tab:aspb}
\begin{tblr}{c | c | c c c c c c}
\toprule
  Pseudo-Box Size & AP & $\mathrm{AP}_{10}$ & $\mathrm{AP}_{30}$ & $\mathrm{AP}_{50}$ & $\mathrm{AP}_{70}$ & $\mathrm{AP}_{90}$ \\
\midrule
  % 20220721_095820.log
  None & 71.9 & 88.9 & 86.8 & 77.7 & 66.2 & 32.9 \\
  % untitled markdown.md
  $p=s$ & 75.6 & 89.2 & 86.3 & 80.3 & \textbf{74.9} & \textbf{37.7} \\
  % 20220721_153436.log
  $p=1.5s$ & \textbf{77.6} & \textbf{93.6} & \textbf{92.6} & \textbf{83.7} & 73.9 & 35.5 \\
  % 20220721_145411.log
  $p=2s$ & 71.2 & 90.7 & 83.2 & 77.2 & 67.5 & 27.0 \\
\bottomrule
\end{tblr}
\end{table}

\subsubsection{Reasonableness of Top-Down Refinement}

% Several studies have demonstrated the effectiveness of top-down modulation between the layers of a feature pyramid \cite{GRSL22CrossConnected}. In contrast to these works \cite{WACV21ACM,TGRS2021ALCNet}, this paper implements top-down modulation at the detection head instead of the neck, allowing us to directly assess the effect of predicting on feature maps of different scales on detection results.
We further investigate the reasons for the performance gains of top-down modulation in our OSCAR head. Specifically, we examine two questions: (1) Is it regression or classification that requires higher resolution feature maps more? (2) How do feature map resolutions affect classification and regression separately?

\begin{figure*}[htbp]
  \centering
  \subfloat[The architecture of SimpleGrid]{
    \includegraphics[height=.18\textwidth]{
      "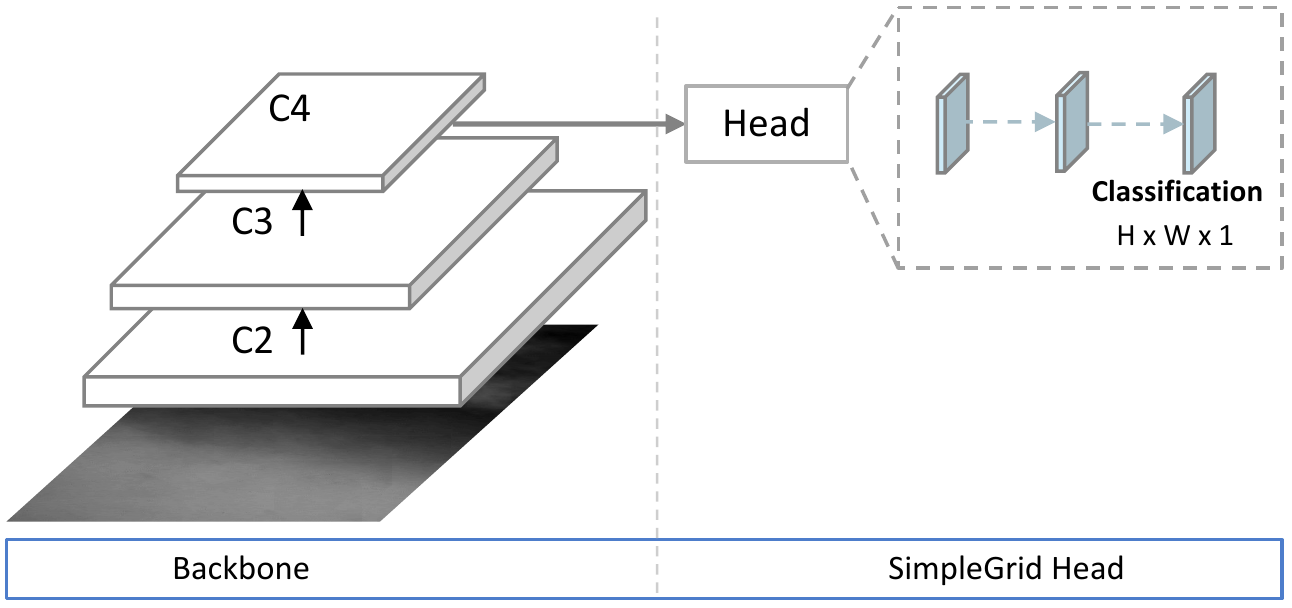"}
      % \label{subfig:plot-neg-pos-ratio}
  }
  \subfloat[SimpleGrid Label Assignment]{
    \includegraphics[height=.22\textwidth]{
      "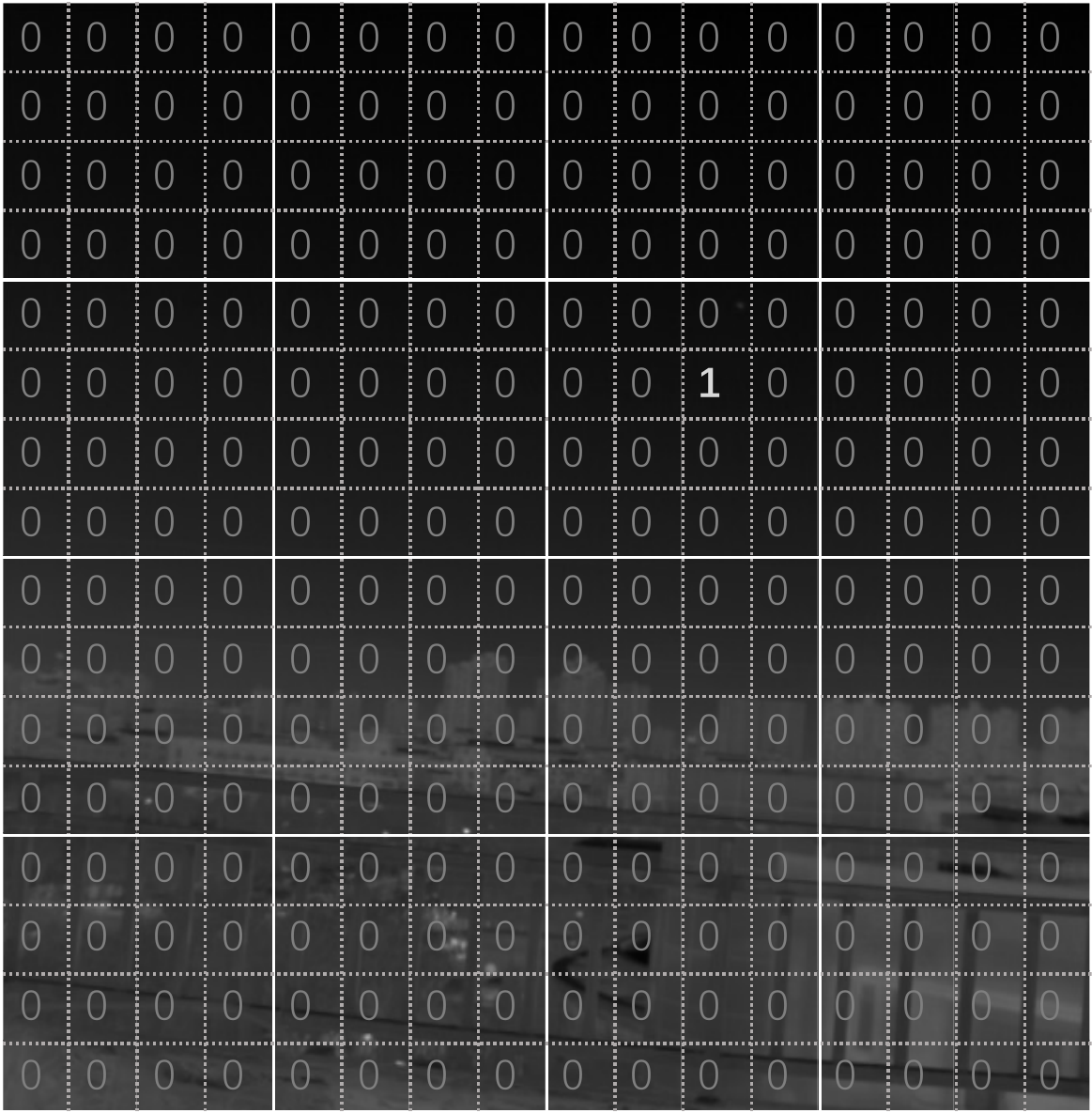"}
      % \label{subfig:plot-neg-pos-ratio}
  }
  \subfloat[Oracle Localization vs BBox Regression]{
    \includegraphics[height=.22\textwidth]{
      "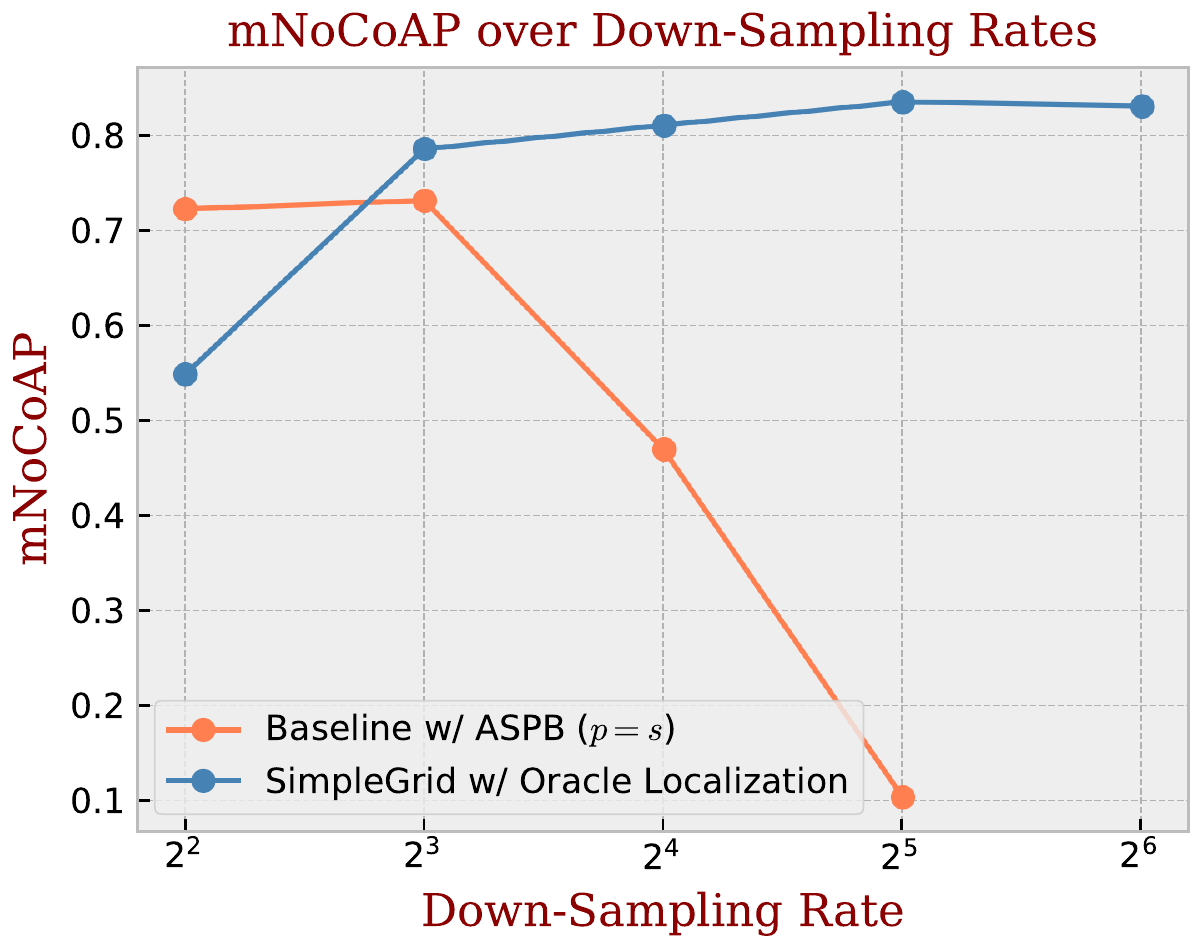"}
      \label{subfig:plot-compute-load}
  }
  \caption{
    Illustration for the ablation study of the reasonableness of our top-down refinement.
    In (a), we show the network architecture of the SimpleGrid, which is equipped with oracle localization to individually inspect the classification performance over various feature map sizes.
    In (b), we show the label assignment for the SimpleGrid, where a patch is considered a positive sample if the target centroid lies within it.
    In (c), we compare the performance of SimpleGrid and the baseline, demonstrating that for infrared small target detection, a larger feature map improves the accuracy of bounding box regression rather than the accuracy of classification. In contrast, the latter faces a more serious class imbalance problem as the feature map becomes larger.}
  \label{fig:reasonable}
  \vspace{-1\baselineskip}
\end{figure*}

To study how feature map resolution affects the classification sub-task, we employ a simple network called SimpleGrid, whose architecture is shown in \cref{fig:reasonable}(a). We endow it with the ability of oracle localization so that the performance can still be expressed in terms of mNoCoAP. Specifically, we extract the groundtruth coordinates and use them to identify the target centroid as long as the feature points are accurately classified.
The label assignment scheme for the SimpleGrid is shown in \cref{fig:reasonable}(b). We divide an input image into a regular grid of equally sized patches. A patch is considered a positive sample if the target centroid lies within it. We keep the input image size constant, but use different down-sampling rates for the network to obtain the prediction results with different feature map sizes, as shown in \cref{tab:downsample}.

\setlength{\tabcolsep}{4pt}
\begin{table}[htbp]
\centering
\caption{Backbone down-sampling schemes on feature map resolutions.}
\label{tab:downsample}
\begin{tblr}{c c c c c c c c}
\toprule
Down-sampling & 4 & 8 & 16 & 32 & 64 \\
\midrule
Stem & $\downarrow 2$ & $\downarrow 2$ & $\downarrow 2$ & $\downarrow 2$ & $\downarrow 2$ \\
Max-Pooling & $\downarrow 2$ & $\downarrow 2$ & $\downarrow 2$ & $\downarrow 2$ & $\downarrow 2$ \\
Stage-1 &  & & & & $\downarrow 2$ \\
Stage-2 &  & $\downarrow 2$ & $\downarrow 2$ & $\downarrow 2$ & $\downarrow 2$ \\
Stage-3 &  & & $\downarrow 2$ & $\downarrow 2$ & $\downarrow 2$ \\
Stage-4 &  & & & $\downarrow 2$ & $\downarrow 2$ \\
\bottomrule
\end{tblr}
\end{table}

The results are shown in \cref{fig:reasonable}(c).
As can be seen, with perfect oracle localization, the performance of SimpleGrid, which only needs to classify feature points, increases as the feature map resolution decreases. In contrast, the performance of the baseline, which performs both bounding box regression and classification, decreases as the down-sampling rate increases. This answers our previous question: it is bounding box regression, not classification, that requires high-resolution feature maps. In fact, the classification accuracy drops as the feature map size increases due to severe class imbalance.

Based on these observations, we can draw the following conclusions. First, low-resolution feature maps are sufficient for preserving infrared small targets. Therefore, if the goal is to determine the presence or absence of a target in a certain region rather than its precise location, low-resolution feature maps are more appropriate. Second, since the high-level layers can classify feature points more accurately, they can be used to modulate the features of the lower layers, which is the source of the performance gains from our top-down modulation.

\subsubsection{Necessity of Joint Classification and Regression}

In our previous ablation study, we adopted a top-down modulation approach to jointly predict the target category and location.
This raises the question of whether the classification and bounding box regression sub-tasks can be directly conducted on different layers, using coarse high-level feature maps for classification and refined low-level feature maps for regression.
To investigate this, we implemented a decoupled head, as shown in \cref{fig:ablation-arch} (b), and compared it to the anchor-free baseline. This allowed us to evaluate the necessity of joint classification and regression.

\setlength{\tabcolsep}{4pt}
\begin{table}[htbp]
\centering
\caption{Ablation study for joint classification and regression.}
\label{tab:decoupled}
\begin{tblr}{c | c | c c c c c}
\toprule
  & AP & $\mathrm{AP}_{10}$ & $\mathrm{AP}_{30}$ & $\mathrm{AP}_{50}$ & $\mathrm{AP}_{70}$ & $\mathrm{AP}_{90}$ \\
\midrule
  Baseline      & \textbf{75.6} & \textbf{89.2} & \textbf{86.3} & \textbf{80.3} & \textbf{74.9} & \textbf{37.7} \\
  DecoupledHead & 55.7 & 81.9 & 66.0 & 59.9 & 48.7 & 14.5 \\
\bottomrule
\end{tblr}
\end{table}
% \vspace{-1\baselineskip}

In \cref{tab:decoupled}, we can see that decoupling the classification and localization sub-tasks into different layers leads to a significant drop in the final detection performance, from 75.6\% to 55.7\%.
We believe that this decline is due to the one-to-many relationship between high and low-level feature points, which allows low-quality bounding boxes predicted by redundant feature points to rank highly, thus significantly reducing the final average precision.
This sharp decline highlights the importance of jointly predicting the target category and location.

\subsubsection{Impact of Top-Down Cascade Refinement}

The proposed OSCAR model has two major differences compared to other cascaded detectors: top-down score modulation (TDSM) and top-down cascaded regression (TDCR). TDSM views the high-level predictions as coarse estimates of the infrared small targets in the low-level refinement head, while TDCR uses a multi-stage bounding box regression to predict the final bounding box more accurately than the single-stage method.

In this part, we investigate the impact of TDSM and TDCR on the detection performance of OSCAR. The results are shown in \cref{tab:oscar}. As can be seen from the table, both TDCR and TDSM alone improve the detection performance of the baseline. Combining the two gives the best performance, improving the baseline from 75.6\% to 79.2\%.
We attribute the performance gain from TDSM to the fact that more accurate classification information at higher levels suppresses the scores of target-like background interferences.
% , causing them to be eliminated or ranked behind in the AP calculation.
The performance gain from TDCR comes from its ability to predict the final bounding box more accurately using a multi-stage approach.

\setlength{\tabcolsep}{4pt}
\begin{table}[htbp]
\centering
\caption{Ablation study for Impact of Top-Down Cascade Refinement.}
\label{tab:oscar}
\begin{tblr}{c c | c | c c c c c}
\toprule
TDSM & TDCR & AP & $\mathrm{AP}_{10}$ & $\mathrm{AP}_{30}$ & $\mathrm{AP}_{50}$ & $\mathrm{AP}_{70}$ & $\mathrm{AP}_{90}$ \\
\midrule
  % untitled markdown.md
  \xmark & \xmark & 75.6 & 89.2 & 86.3 & 80.3 & 74.9 & 37.7 \\
  $\checkmark$ & \xmark & 78.5 & 88.9 & 86.7 & 84.2 & 71.9 & \textbf{54.0} \\
  \xmark & $\checkmark$ & 78.2 & 93.1 & \textbf{89.4} & 84.7 & \textbf{78.4} & 36.2 \\
  $\checkmark$ & $\checkmark$ & \textbf{79.2} & \textbf{94.8} & 89.3 & \textbf{86.3} & 74.4 & 43.2 \\
\bottomrule
\end{tblr}
\end{table}

\subsubsection{With or Without Normalized Contrast}

In this part, we investigate the impact of the normalized contrast branch. As mentioned previously, this branch is designed to suppress the low-quality detected bounding boxes produced by feature points that are far from the infrared small targets. These false alarms are introduced by the pseudo-boxes in our ASPB label assignment scheme.
The results of this investigation are shown in \cref{tab:noco}. In this table, "None" indicates that no normalized contrast is used, "NoCo" denotes that the normalized contrast is used, "NoCo$^\dag$" stands for that the normalized contrast is multiplied with the interpolated classification scores from the soft proposal head as the final score, and "Centerness$^\dag$" means that the centerness is used instead of the normalized contrast.

As can be seen from the table, the augmented branch effectively suppresses the false alarms introduced by the pseudo-boxes, which improves the performance from 79.2\% to 80.3\%. Additionally, the normalized contrast branch achieves a higher score than the centerness, which is due to the fact that in our mNoCoAP metric, the centroid of an infrared small target is annotated with the pixel of the highest signal-to-clutter ratio, while the centerness is merely a sub-optimal proxy since the center pixel does not necessarily have the highest local contrast.

\setlength{\tabcolsep}{4pt}
\begin{table}[htbp]
\centering
\caption{Ablation study for the proposed normalized contrast branch.}
\label{tab:noco}
\begin{tblr}{c | c | c c c c c}
\toprule
  & AP & $\mathrm{AP}_{10}$ & $\mathrm{AP}_{30}$ & $\mathrm{AP}_{50}$ & $\mathrm{AP}_{70}$ & $\mathrm{AP}_{90}$ \\
\midrule
  None              & 79.2          & \textbf{94.8} & 89.3          & \textbf{86.3} & 74.4          & 43.2 \\
  NoCo              & 79.9          & 90.8          & 89.2          & 84.6          & 77.3          & 49.1 \\
  Centerness$^\dag$ & 79.7          & 92.9          & \textbf{90.1} & 81.9          & 75.5          & \textbf{50.9} \\
  NoCo$^\dag$       & \textbf{80.3} & 91.7          & 89.5          & 85.9          & \textbf{77.4} & 47.9 \\
\bottomrule
\end{tblr}
\end{table}

\subsubsection{Choice of backbone depth and prediction level}

In generic computer vision, the detection performance is usually improved gradually as the depth of the network increases. However, this is not the case for infrared small targets. The features that highlight small objects tend to be extinguished because of the down-sampling operations in the backbone, which can result in the contamination of the features of infrared small targets by noise in the background in deep networks.
In addition, different prediction levels result in very different computational effort. For example, the computational load of our OSCAR head on the P2 layer is 300\% more than predicting on the P3 layer. Therefore, it is necessary to explore the optimal network depth and prediction level to balance the performance and efficiency.

In \cref{tab:depth}, we provide the mNoCoAP values and GFLOPS of our OSCAR network on different backbones and prediction levels.
Here, $C$ is the output channel number of the feature pyramid and $s$ is the feature map stride of the refinement layer.
As shown in the table, the detection performance of the network with the same parameter settings decreases as the network depth increases. Therefore, the ResNet-18 is already sufficient for the infrared small target detection task, and it is more important to focus on how to more accurately regress targets than to use a more powerful backbone to improve detection performance.
When we predict the target at a shallower layer, the detection performance is better, but at the cost of a 2-3 times higher computational effort. Finally, we can significantly reduce the computation on top of the P2 layer by reducing the number of channels in the head. As shown in the table, by halving the number of channels, we can achieve better performance (83.52\% vs 82.77\%) while maintaining less computation than the P3 layer prediction (25.36 vs 25.55 GFLOPS).

\setlength{\tabcolsep}{4pt}
\begin{table*}[htbp]
\centering
\caption{Ablation study on OSCAR's choice of backbone depth and prediction level}
\label{tab:depth}
% \vspace{-.5\baselineskip}
% \footnotesize
\small
\begin{tabular}{Sc Sc Sc Sc Sc Sc Sc Sc Sc}
\toprule
Backbone & \multicolumn{2}{c}{ResNet-18} & \multicolumn{2}{c}{ResNet-34} & \multicolumn{2}{c}{ResNet-50} &  \multicolumn{2}{c}{ResNet-101} \\
\cmidrule(lr){2-3} \cmidrule(lr){4-5} \cmidrule(lr){6-7} \cmidrule(lr){8-9}
Refining Layer & mNoCoAP   & GFLOPS & mNoCoAP   & GFLOPS & mNoCoAP   & GFLOPS & mNoCoAP   & GFLOPS \\
\midrule
P4 $(C = 256, s = 16)$ & \textbf{74.38} & 14.85  & 72.66 & 24.54  & 71.36 & 28.37  & 70.82 & 47.80 \\
P3 $(C = 256, s = 8)$  & \textbf{82.77} & 25.55  & 82.04 & 35.23  & 80.60 & 39.06  & 80.24 & 58.54 \\
P2 $(C = 256, s = 4)$  & \textbf{85.02} & 68.32  & 82.98 & 78.00  & 82.33 & 81.84  & 82.12 & 101.31 \\
\midrule
P2 $(C = 128, s = 4)$  & \textbf{83.52} & 25.36 & 82.15 & 35.04 & 80.95 & 38.12 & 80.75 & 57.59 \\
\bottomrule
\end{tabular}
% \vspace*{-2.\baselineskip}
\end{table*}

% 网络并不是越深越好, 对于
% 这是因为目标特征容易消失;
% P2 可以获得更好的性能, 但是计算量也更大.

\subsubsection{Choice of Hyper-paramter $\lambda$}
In this part, we study the choice of the hyper-parameter $\lambda$, which is the weight of the loss of the normalized contrast branch in the OSCAR network. As shown in \cref{tab:lambda}, the tested OSCAR networks (with a ResNet-50 backbone and predicting on the P2 head) achieve the highest mNoCoAP value when $\lambda=10^3$. As a result, we choose this value as the default parameter.

\setlength{\tabcolsep}{4pt}
\begin{table}[htbp]
\centering
\caption{Ablation study for Impact of Top-Down Cascade Refinement.}
\label{tab:lambda}
\begin{tblr}{c | c | c c c c c}
\toprule
$\lambda$ & AP & $\mathrm{AP}_{10}$ & $\mathrm{AP}_{30}$ & $\mathrm{AP}_{50}$ & $\mathrm{AP}_{70}$ & $\mathrm{AP}_{90}$ \\
\midrule
  % untitled markdown.md
$10^2$ & 80.2          & 92.4          & 88.5          & 85.5          & \textbf{78.3} & 49.9 \\
$10^3$ & \textbf{80.6} & 93.4          & \textbf{91.6} & 81.9          & 77.3          & \textbf{51.6} \\
$10^4$ & 78.5          & \textbf{94.5} & 88.6          & \textbf{85.7} & 77.3          & 43.9 \\
$10^5$ & 77.2          & 92.9          & 89.1          & 82.1          & 74.0          & 42.6 \\
\bottomrule
\end{tblr}
\end{table}

\subsection{Comparison with State-of-the-Arts}\label{subsec:sota}

In this subsection, we compare our OSCAR model with several model-driven and data-driven methods, including some state-of-the-art deep networks.
The mNoCoAP values and inference times are listed in \cref{tab:sota}.
The inference speed comparison is provided in terms of FPS.
Traditional methods are evaluated on the original size of the images, while deep learning methods are evaluated on resized images with a specific width and height, \textit{e.g.}, $(1000, 600)$.
% We will delve into the following two parts for an in-depth discussion: (1) Comparison of model-driven and data-driven methods, and (2) Comparison among data-driven methods.

\subsubsection{Model-driven versus Data-driven}
It can be seen that data-driven methods outperform in terms of both detection performance and detection speed. This large performance gap shows that the global sparsity and contrast priors no longer hold for model-driven methods when dealing with complicated images, yet data-driven methods can still produce satisfying detection results.
We attribute this performance gain to two main factors. First, data-driven methods use deep networks to extract high-level semantics from the images, which allows them to better distinguish real targets from false alarms that look similar to infrared small targets. Second, data-driven methods have fewer hyper-parameters than model-driven methods, making them more robust to variations in the scene.

In addition, traditional methods, particularly low-rank approaches, require dozens of iterations to make predictions, whereas deep networks can make predictions in a single forward pass. This makes data-driven methods faster and more efficient. Furthermore, the computational complexity of traditional methods grows exponentially with the size of the input image, while data-driven methods maintain a linear relationship with the image size.
As infrared imaging technology advances, infrared images of high resolutions, even higher than $1280 \times 1024$ in SIRST-V2, will become more common in the future. Therefore, data-driven methods should receive more attention from researchers, both in terms of detection performance and detection speed.

\setlength{\tabcolsep}{1pt}
\begin{table*}[htbp]
\centering
\caption{Comparison with other state-of-the-art methods on mNoCoAP.}
\label{tab:sota}
% \vspace{-.5\baselineskip}
% \footnotesize
\small
\begin{tabular}{Sc Sc Sc Sc Sc Sc Sc Sc Sc Sc Sc Sc Sc Sc Sc Sc Sc}
\toprule
\multirow{3}{*}{Methods} & \multicolumn{8}{c}{Model-Driven} & \multicolumn{7}{c}{Data-Driven} \\
\cmidrule(lr){2-9} \cmidrule(lr){10-17}
  & \multicolumn{4}{c}{Local Contrast} & \multicolumn{4}{c}{Low-Rank + Sparse} & \multicolumn{3}{c}{Semantic Seg.} & \multicolumn{5}{c}{BBox Regression} \\
\cmidrule(lr){2-5} \cmidrule(lr){6-9} \cmidrule(lr){10-12} \cmidrule(lr){13-17}
  & FKRW & LCM & MPCM & WLDM & IPI & NIPPS & RIPT & SMSL & ISTDU & ACMNet & ALCNet & FCOS & Faster RCNN & RFLA & QueryDet & OSCAR \\
\midrule
  AP & 27.8 & 20.7 & 32.2 & 11.2 & 37.7 & 33.5 & 29.2 & 20.6 & 65.1 & 61.4 & 66.4 & 73.1 & 73.4 & 78.8 & 80.1 & \textbf{85.2} \\
\midrule
$\mathrm{AP}_{10}$ & 31.9 & 31.1 & 34.4 & 13.4 & 39.6 & 35.5 & 34.6 & 23.1 & 83.3 & 85.5 & 81.1 &  87.7 & 88.7 & 92.5 & 93.0 &  \textbf{93.8} \\
$\mathrm{AP}_{20}$ & 31.6 & 30.3 & 33.8 & 13.1 & 39.6 & 35.4 & 33.8 & 22.7 & 82.6 & 82.5 & 79.5 &  87.0 & 88.7 & 91.5 & 93.0 &  \textbf{93.4} \\
$\mathrm{AP}_{30}$ & 30.5 & 28.6 & 33.5 & 13.1 & 39.0 & 34.8 & 33.3 & 22.2 & 81.3 & 79.5 & 77.5 &  84.7 & 86.7 & 90.2 & 91.5 &  \textbf{93.3} \\
$\mathrm{AP}_{40}$ & 29.8 & 27.4 & 33.3 & 13.1 & 38.3 & 34.2 & 33.3 & 22.2 & 79.2 & 79.5 & 76.1 &  81.0 & 84.4 & 88.1 & 89.1 & \textbf{90.4} \\
$\mathrm{AP}_{50}$ & 28.5 & 23.3 & 33.3 & 12.5 & 38.3 & 34.2 & 32.1 & 21.2 & 76.2 & 70.6 & 72.1 &  77.5 & 80.7 & 84.0 & 86.5 & \textbf{88.8} \\
$\mathrm{AP}_{60}$ & 28.5 & 19.9 & 33.3 & 11.3 & 38.3 & 33.7 & 28.8 & 20.5 & 69.4 & 59.8 & 68.4 &  71.4 & 74.8 & 79.9 & 83.0 & \textbf{86.2} \\
$\mathrm{AP}_{70}$ & 27.2 & 13.2 & 32.2 & 10.1 & 38.3 & 33.0 & 26.1 & 19.5 & 57.4 & 48.4 & 57.1 &  69.6 & 66.8 & 72.8 & 76.6 & \textbf{82.3} \\
$\mathrm{AP}_{80}$ & 26.6 &  7.8 & 31.2 &  8.1 & 36.2 & 31.3 & 23.5 & 19.0 & 39.9 & 31.1 & 50.4 &  59.8 & 54.1 & 63.7 & 63.6 & \textbf{76.9} \\
$\mathrm{AP}_{90}$ & 15.2 & 4.8  & 24.9 & 6.0  & 32.0 & 29.5 & 17.2 & 17.3 & 16.3 & 16.0 & 35.2 &  39.4 & 35.4 & 46.6 & 45.2 & \textbf{60.2} \\
\midrule
% Runtime/s & 1.913 & 0.126 & 0.899 & 7.933 & 18.30 & 18.64 & 9.273 & 2.667 & 0.052 & 0.034 & 0.381 & 0.035 & 0.235 & 0.237 & 0.084 & 0.037 \\
FPS & 0.52 & 7.93 & 1.11  & 0.12  & 0.05 & 0.05 & 0.11  & 0.37  & 19.23  & 29.41  & 2.62 & 28.57  & 4.25 & 4.21 & 11.90 & \textbf{34.6} \\
\bottomrule
\end{tabular}
% \vspace*{-2.\baselineskip}
\end{table*}

\subsubsection{Among Data-Driven Methods}
Then we compare the proposed OSCAR method with other deep learning-based approaches. Our results show that:
(1)~Semantic segmentation approaches are outperformed by all bounding box regression methods. We believe the reason for this lies in the impact of uncertainty in the labels of pixels caused by the dispersion effect of long-distance imaging, which makes up the vast majority of positive samples and predominates the training loss. Since the objective of this task is to locate the centroids of infrared small targets rather than their entire contours, bounding box regression is a more reasonable detection pipeline for infrared small targets.
(2)~Our proposed OSCAR method outperforms FCOS by 7.2\%. This performance gain demonstrates the effectiveness of our approach in achieving the primary goal of this paper, which is to improve detection results by regressing the target bounding box coordinates in a two-step cascade.
(3)~Compared to the two-stage Faster R-CNN, which also regresses the final coordinates based on initial guesses, our proposed OSCAR method achieves a much better result. We believe this is because OSCAR avoids using anchor-related hyper-parameters, which are known to be sensitive to the final detection performance. Additionally, Faster R-CNN assigns multiple samples per feature point, whereas OSCAR only assigns one, which makes the class imbalance more severe for Faster R-CNN and leads to inferior classification accuracy.

In addition to outperforming FCOS, our proposed OSCAR method also outperforms both RFLA and QueryDet.
It is important to note that, although both RFLA and QueryDet are designed for small objects, the small objects they assume still have a scale gap with the infrared small targets used in this study.
RFLA is generally designed for objects with sizes ranging from $16 \times 16$ to $32 \times 32$, while OSCAR is typically used for objects smaller than $12 \times 12$, mostly in the range of $3 \times 3$ to $5 \times 5$. Therefore, the poorer performance of RFLA is predictable because it predicts small targets in the P3 head at a stride of $8$, which is too large for detecting infrared small targets.
As for QueryDet, our OSCAR method differs in the way it utilizes more information from the high-level head. We believe this is the reason why OSCAR achieves better performance than QueryDet. In QueryDet, only the existence of small objects is passed down from the P3 head to the P2 head, whereas OSCAR fully uses the predictions (both classification and regressed bounding boxes) produced by the coarse head.

\begin{figure*}[htbp]
  \centering
  \includegraphics[width=.985\textwidth]{"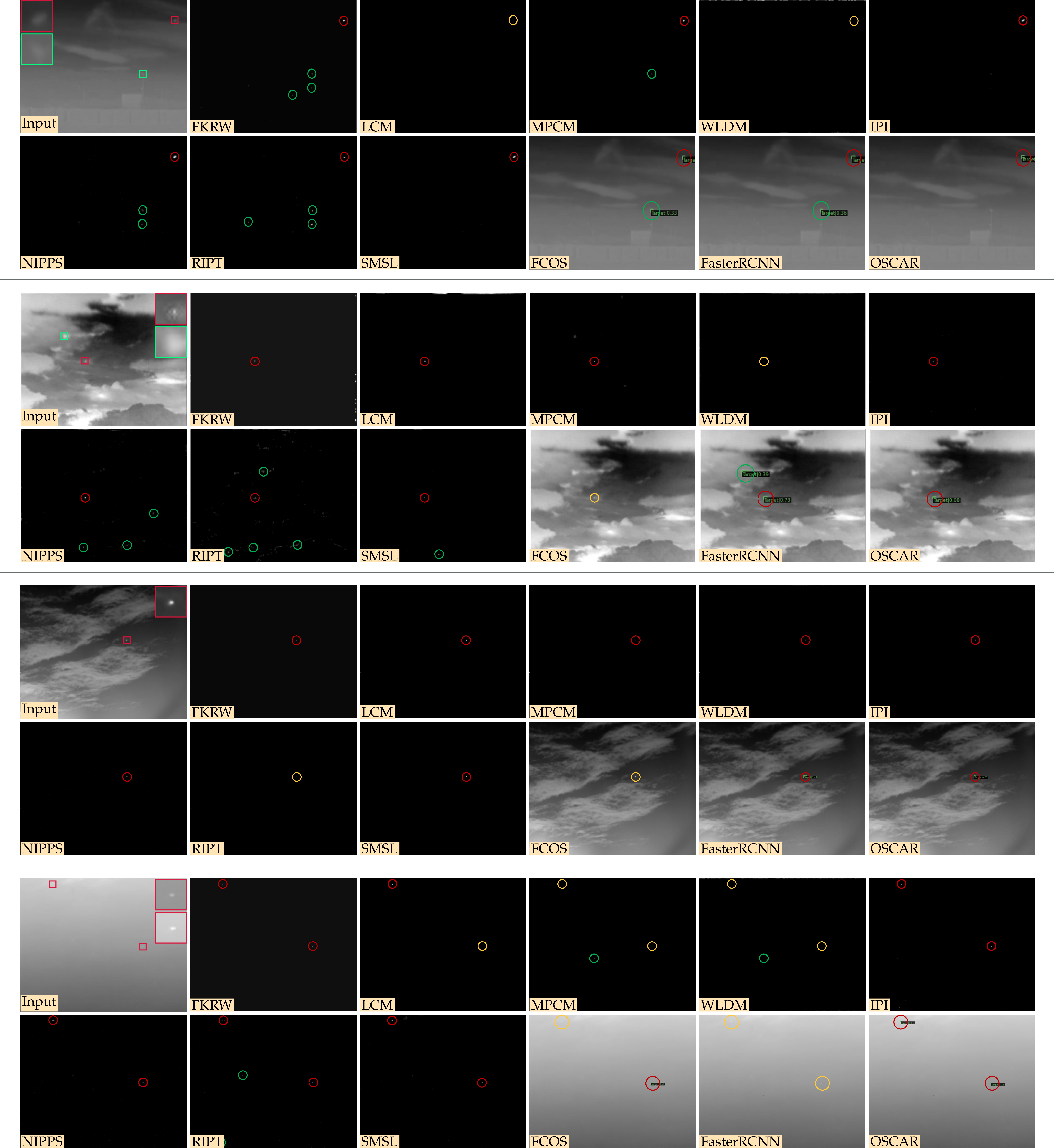"}
  \caption{
    Visualization of predicted results by tested methods on four real images containing infrared small targets.
    In the input images, {\color{WildStrawberry} \textbf{real targets}} are indicated by the {\color{WildStrawberry} \textbf{red boxes}}, whereas {\color{YellowGreen} \textbf{background distractors}} (target-like background components) are indicated by the {\color{YellowGreen} \textbf{green boxes}}.
    In the rest detection results, {\color{WildStrawberry} \textbf{real circles}} denote the {\color{WildStrawberry} \textbf{True Positive (TP)}} detections;
    {\color{YellowGreen} \textbf{green circles}} denote the {\color{YellowGreen} \textbf{False Positive (FP)}} detections;
    {\color{Dandelion} \textbf{yellow circles}} denote the {\color{Dandelion} \textbf{False Negative (FN) }} detections.
    The better the detection performance is indicated by the greater number of red circles and the fewer green and yellow circles.
    In general, low-rank methods outperform local contrast methods in terms of missed detections, but are still outperformed by deep learning approaches in suppressing false alarms.
    It is worth noting that FCOS tends to miss small targets, while the proposed OSCAR is able to detect them well.
    We believe this improvement in performance is due to our pseudo-box based label assignment, which addresses the issue of mislabeling targets.
    For best viewing, the image should be zoomed in on a computer screen.
    }
  \label{fig:vis-target}
\end{figure*}

\begin{figure*}[htbp]
  \centering
  \includegraphics[width=.985\textwidth]{"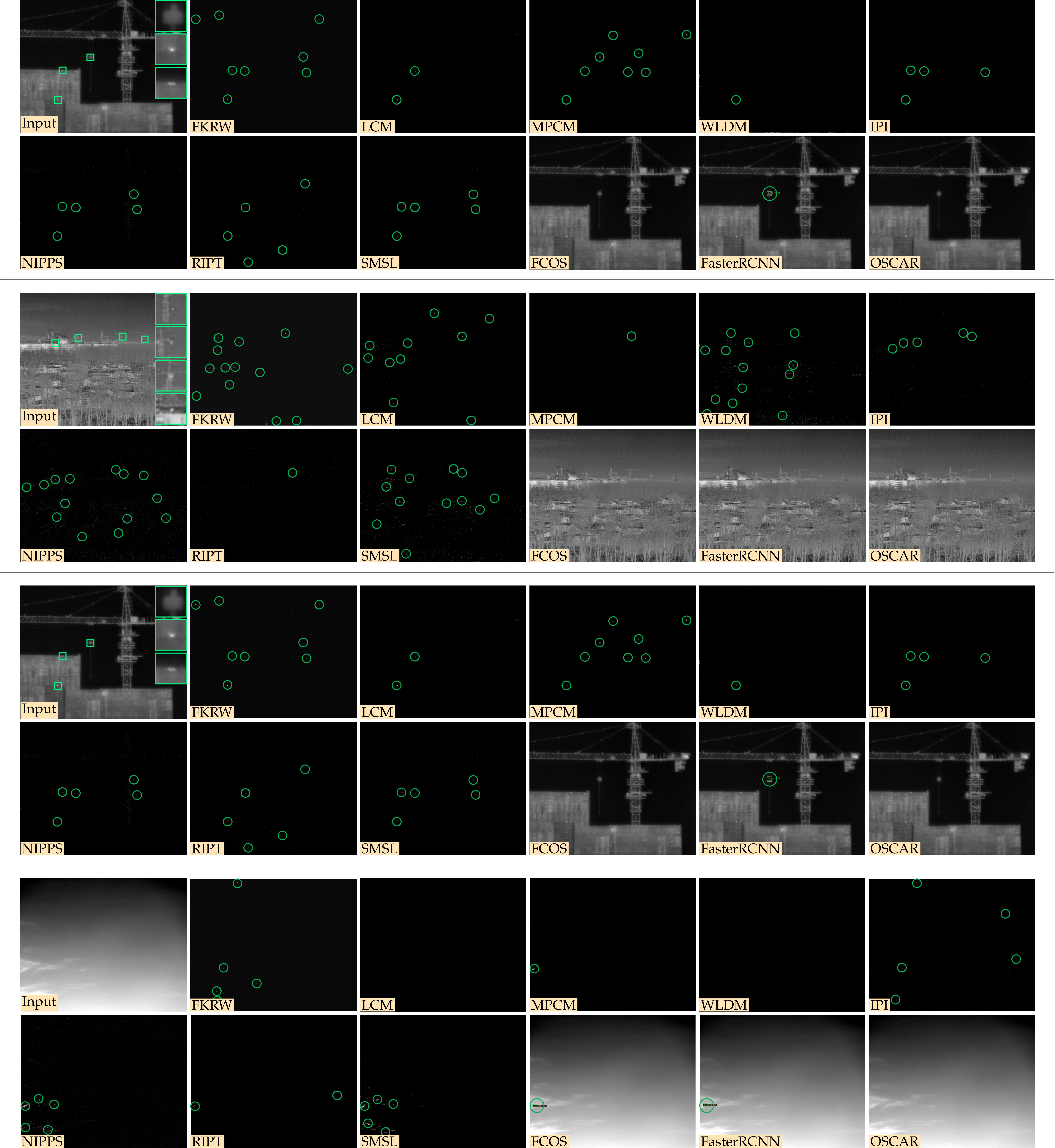"}
  \caption{
    Visualization of predicted results by tested methods on four background images without any real target.
    The goal of this experiment set is to evaluate the ability of the methods to suppress false alarms.
    % This set of experiments focuses on the ability of the method to suppress false alarms.
    % \textbf{Green}: False Positive (FP) Detections.
    In the input images, {\color{YellowGreen} \textbf{background distractors}} (target-like background components) are indicated by the {\color{YellowGreen} \textbf{green boxes}}.
    In the rest detection results, {\color{YellowGreen} \textbf{green circles}} denote the {\color{YellowGreen} \textbf{False Positive (FP)}} detections.
    % The less green circles, the better the false alarm suppression performance.
    As can be seen, traditional methods produce a very high number of false alarms, while deep learning methods are able to extract the semantics of the images and generate far fewer false alarms. Among the deep learning methods, the OSCAR method proposed in this paper performs the best, not generating any false alarms for these four images. These results demonstrate the effectiveness of OSCAR in suppressing false alarms and accurately detecting infrared small targets.}
  \label{fig:vis-background}
\end{figure*}

\subsection{Visual Analysis}

To showcase the benefits of the proposed OSCAR, we provide visualizations of the experimental results in two groups, \cref{fig:vis-target} and \cref{fig:vis-background}, depending on whether the test images contain real targets.
We use {\color{WildStrawberry} \textbf{real circles}} to denote {\color{WildStrawberry} \textbf{True Positive (TP)}} detections, {\color{YellowGreen} \textbf{green circles}} for {\color{YellowGreen} \textbf{False Positive (FP)}} detections, and {\color{Dandelion} \textbf{yellow circles}} for {\color{Dandelion} \textbf{False Negative (FN) }} detections.
% It is intuitive that t
The better the performance, the more red circles there are and the fewer green and yellow circles there are.

From \cref{fig:vis-target}, it can be seen that the traditional methods, such as FKRW, IPI, NIPPS, and SMSL, are able to detect the targets in the images, but may generate false alarms.
In particular, the IPI method performed better because it did not generate false alarms, which is consistent with its good performance among traditional methods in the quantitative metrics, as shown in \cref{tab:sota}.
It can be concluded that low-rank methods typically outperform local contrast methods in terms of detection accuracy.
Among the deep learning approaches, OSCAR is able to detect real targets while suppressing false alarms.
It should be noted that FCOS often misses the tiny real targets in the chosen test images.
This is likely because FCOS's label assignment ignores real targets whose bounding boxes do not include any feature points, preventing the smallest scale targets from being trained.
In contrast, our proposed OSCAR solves this problem by using the pseudo-box for label assignment and performs better in detecting very small scale targets.
Overall, this visualization of the results highlights the importance of semantic understanding and careful label assignment for accurate infrared small target detection.

As shown in \cref{fig:vis-background}, the second group of input images primarily consist of urban scenes, which are newly added in the SIRST-V2 dataset and contain many cranes, street lights, and other non-target background distractions that are similar in scale to the infrared small targets. 
This set of experiments focuses on the ability of different methods to suppress false alarms using semantic understanding of the images. 
The traditional methods shown in \cref{fig:vis-background} all produce a high number of false alarms, because in their models the detection of infrared small targets is simplified to the detection of salient regions of the image that match the scale of the targets.
However, in complex backgrounds not everything that fits the model's appearance prior is a real target, and it is likely to be a background distractor. 
In contrast, deep learning methods can extract the semantics of the images to produce significantly fewer false alarms. 
The performance gap between traditional and deep learning methods is largely due to their ability to suppress false alarms.
Additionally, OSCAR outperforms FCOS in terms of not generating any false alarms for these four images.
We believe this is due to the usage of normalized contrast branch, which helps to suppress false alarms.

In conclusion, our experiments demonstrate that the OSCAR network is a promising approach for detecting infrared small targets. It achieves superior performance on the SIRST-V2 dataset and is efficient in terms of both detection performance and inference speed. We hope that the proposed OSCAR network will serve as a reliable baseline for future research in this area.
% !TEX root = ../main.tex

\section{Conclusion}    \label{sec:conclusion}

We have proposed a comprehensive solution to the challenges of infrared small target detection, including a new research benchmark consisting of the SIRST-V2 dataset, the normalized contrast evaluation metric, and the DeepInfrared toolkit.
Furthermore, our proposed one-stage cascade refinement network outperforms existing methods through three improvements, namely, an all-scale pseudo-box-based label assignment scheme, the use of the high-level head as soft proposals for the low-level refinement head, and a normalized contrast branch for better localization quality.
Extensive experiments and analyses on the SIRST-V2 dataset further validate the effectiveness and efficiency of OSCAR. We hope it can serve as a simple yet effective baseline for the community.

\bibliographystyle{IEEEtran}
% \bibliography{../../ref-bib/all-refs.bib}
\bibliography{./all-refs.bib}

\end{document}